\documentclass[final,5p,times,twocolumn]{elsarticle}

\usepackage{amssymb}
\usepackage{amsmath,amsfonts}
\usepackage{algorithmic}
\usepackage{algorithm}
\usepackage{array}
\usepackage[caption=false,font=normalsize,labelfont=sf,textfont=sf]{subfig}
\usepackage{textcomp}
\usepackage{stfloats}
\usepackage{natbib}
\setcitestyle{numbers,square}
\usepackage{url}
\usepackage{bm}
\usepackage{multirow}
\usepackage{CJKutf8}
\usepackage[table,xcdraw]{xcolor}
\usepackage{color}
\usepackage{graphicx}
\usepackage{tcolorbox}
\usepackage{booktabs}
\usepackage{makecell}
\usepackage{ulem}

\usepackage{tcolorbox}
\newtcolorbox{hypobox}[2][]{%
  colback=white,
  colframe=gray!50!black, 
  colbacktitle=gray!15,    
  coltitle=black,          
  fonttitle=\bfseries,     
  title={#2},              
  sharp corners=south,     
  rounded corners=north,   
  arc=3mm,                 
  boxrule=0.5pt,           
  left=4pt, right=4pt, top=4pt, bottom=4pt, 
  #1
}

\journal{Knowledge-Based Systems}

\begin{document}

\begin{frontmatter}

\title{Overcoming Environmental Meta-Stationarity in MARL via Adaptive Curriculum and Counterfactual Group Advantage}

\author{Weiqiang Jin\fnref{1}}
\ead{weiqiangjin@stu.xjtu.edu.cn}

\author{Yang Liu\fnref{1}}
\ead{liu0725@stu.xjtu.edu.cn}

\author{Shixiang Tang\fnref{1}}
\ead{shixiangtang@stu.xjtu.edu.cn}

\author{Jinhu Qi\fnref{2}}
\ead{jhqi25@cse.cuhk.edu.hk}

\author{Wentao Zhang\fnref{2}}
\ead{p2522808@mpu.edu.mo}

\author{Junli Wang\fnref{3}}
\ead{edwardwang733@gmail.com}

\author{Biao Zhao\corref{cor1}\fnref{1}}
\ead{biaozhao@xjtu.edu.cn}

\author{Hongyang Du\fnref{4}}
\ead{duhy@eee.hku.hk}

\cortext[cor1]{Corresponding author: Biao Zhao.}

\affiliation[1]{organization={School of Information and Communications Engineering, Xi'an Jiaotong University},
            addressline={Innovation Harbour},
            city={Xi'an},
            postcode={710049},
            state={Shaanxi},
            country={China}}

\affiliation[2]{organization={Department of Computer Science and Engineering, The Chinese University of Hong Kong},
            city={Hong Kong},
            postcode={999077},
            country={Hong Kong SAR, China}}

\affiliation[3]{organization={School of Computer Science and Technology, University of Science and Technology of China},
            addressline={No.~96 JinZhai Road, Baohe District},
            city={Hefei},
            postcode={230026},
            state={Anhui},
            country={China}}

\affiliation[4]{organization={Department of Electrical and Electronic Engineering, The University of Hong Kong},
            city={Hong Kong},
            postcode={999077},
            country={Hong Kong SAR, China}}

\begin{abstract}
Multi-agent reinforcement learning (MARL) has reached competitive performance on cooperative tasks against scripted adversaries, yet most methods train agents at a single fixed difficulty throughout the entire run.
We term this static-difficulty regime \textit{environmental meta-stationarity} and show that it caps policy generalization and steers learning toward shallow local optima.
To break this regime, we propose \textit{CL-MARL}, a dynamic curriculum learning framework that adapts opponent strength online from win-rate signals, advancing or regressing the task as agents master it.
Its scheduler, \textit{FlexDiff}, fuses momentum-based trend estimation with sliding-window dual-curve monitoring of training and evaluation returns, yielding stable difficulty transitions without manual tuning.
Because a moving curriculum amplifies non-stationarity and sparsifies global rewards, we introduce the \textit{Counterfactual Group Relative Policy Advantage} (\textit{CGRPA}), which extends GRPO-style group-relative optimization with counterfactual baselines to disentangle each agent's contribution under shifting team dynamics.
On the StarCraft Multi-Agent Challenge (SMAC), CL-MARL attains a \textbf{40\%} mean win rate on the super-hard maps with an average episode return of \textbf{17.85}, exceeding the QMIX, OW-QMIX, DER, EMC, and MARR baselines by \textbf{+2.94} on average, while reaching its peak win rate roughly \textbf{$1.28\times$} faster on \texttt{8m\_vs\_9m} and \textbf{$1.42\times$} faster on \texttt{3s5z\_vs\_3s6z} than the strongest baseline.
The implementation is publicly available at \url{https://github.com/NICE-HKU/CL2MARL-SMAC}.
\end{abstract}
 
 \begin{graphicalabstract}
     \includegraphics[width=2.0\linewidth]{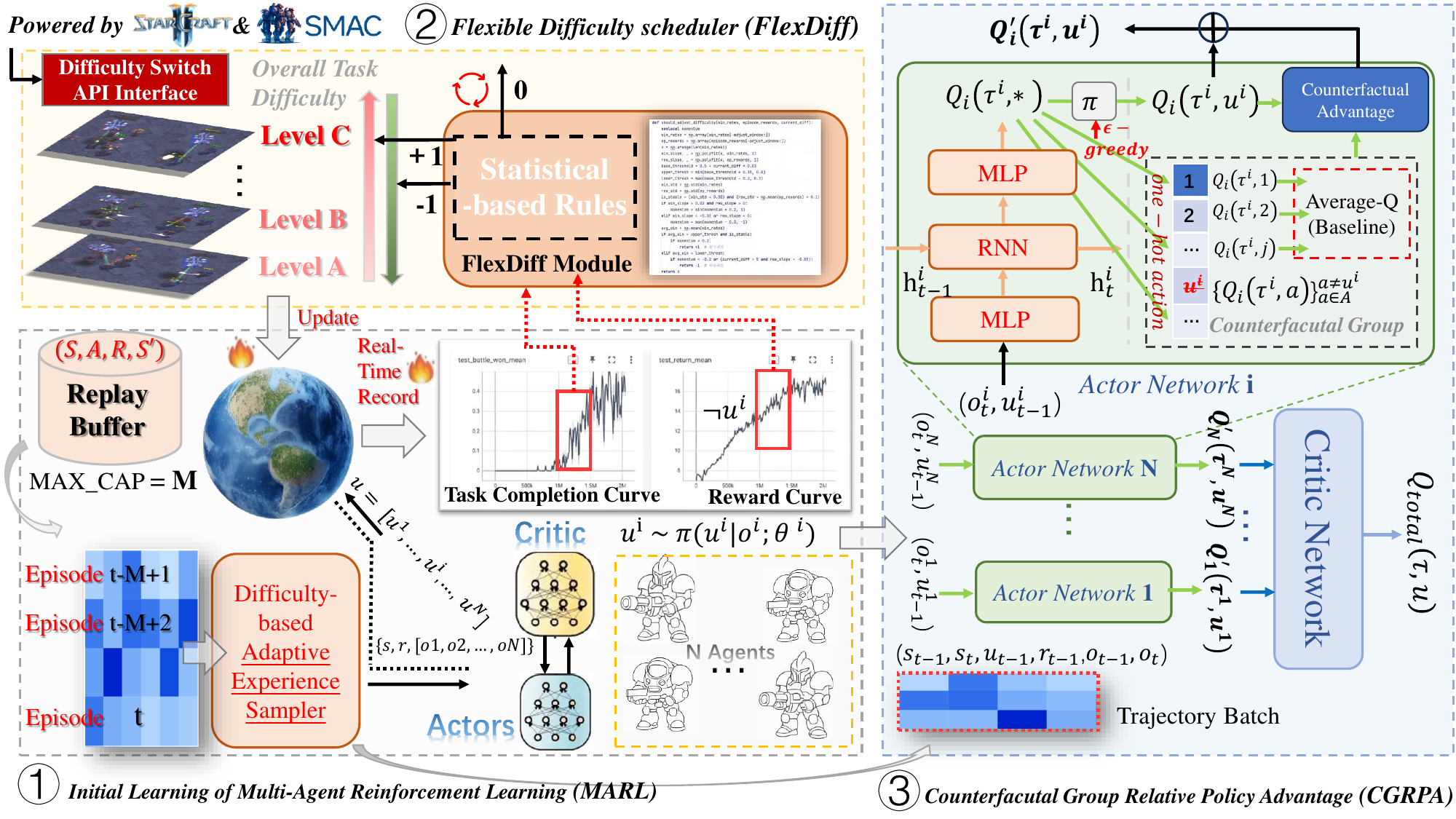}
 \end{graphicalabstract}


\begin{highlights}
\item We identify environmental meta-stationarity, the fixed-difficulty trap in MARL.
\item CL-MARL is a dynamic curriculum that adapts opponent strength from win-rate signals.
\item FlexDiff schedules difficulty via momentum-driven sliding-window dual-curve monitoring.
\item CGRPA fuses GRPO group-relative optimization with COMA counterfactual reasoning.
\item Reaches 40\% on SMAC super-hard maps; +30 pts and $1.42\times$ faster vs.\ baselines.
\end{highlights}


\end{frontmatter}


\section{Introduction}
\label{sec:introduction}
In real-world artificial intelligence (AI) applications, complex problems can be formulated as multi-agent cooperative decision-making tasks, where multiple intelligent agents collaborate to achieve shared objectives \cite{NING2024Survey}. 
Representative examples include traffic signal control, autonomous driving systems, communication resource allocation, and military simulation scenarios \cite{Pamul2023survey,Zhang2024traffic,Zhu2024Survey}. 
Multi-agent reinforcement learning (MARL) has emerged as an effective framework for addressing these problems by enabling agents to learn coordinated strategies through interaction
\cite{Xu9904958SurveyDRL}.
Notably, recent advancements in MARL have led to groundbreaking achievements of multi-agent cooperative tasks, 
with a variety of novel algorithms achieving state-of-the-art (SOTA) results in benchmark environments such as
StarCraft II micromanagement\footnote{PySC2 - StarCraft II: \url{https://github.com/google-deepmind/pysc2}} \cite{ellis2023smacv2} and Google Gfootball \cite{Kurach2020gfootball}.  

\begin{figure*}[ht]
\centering
\includegraphics[width=1.0\linewidth]{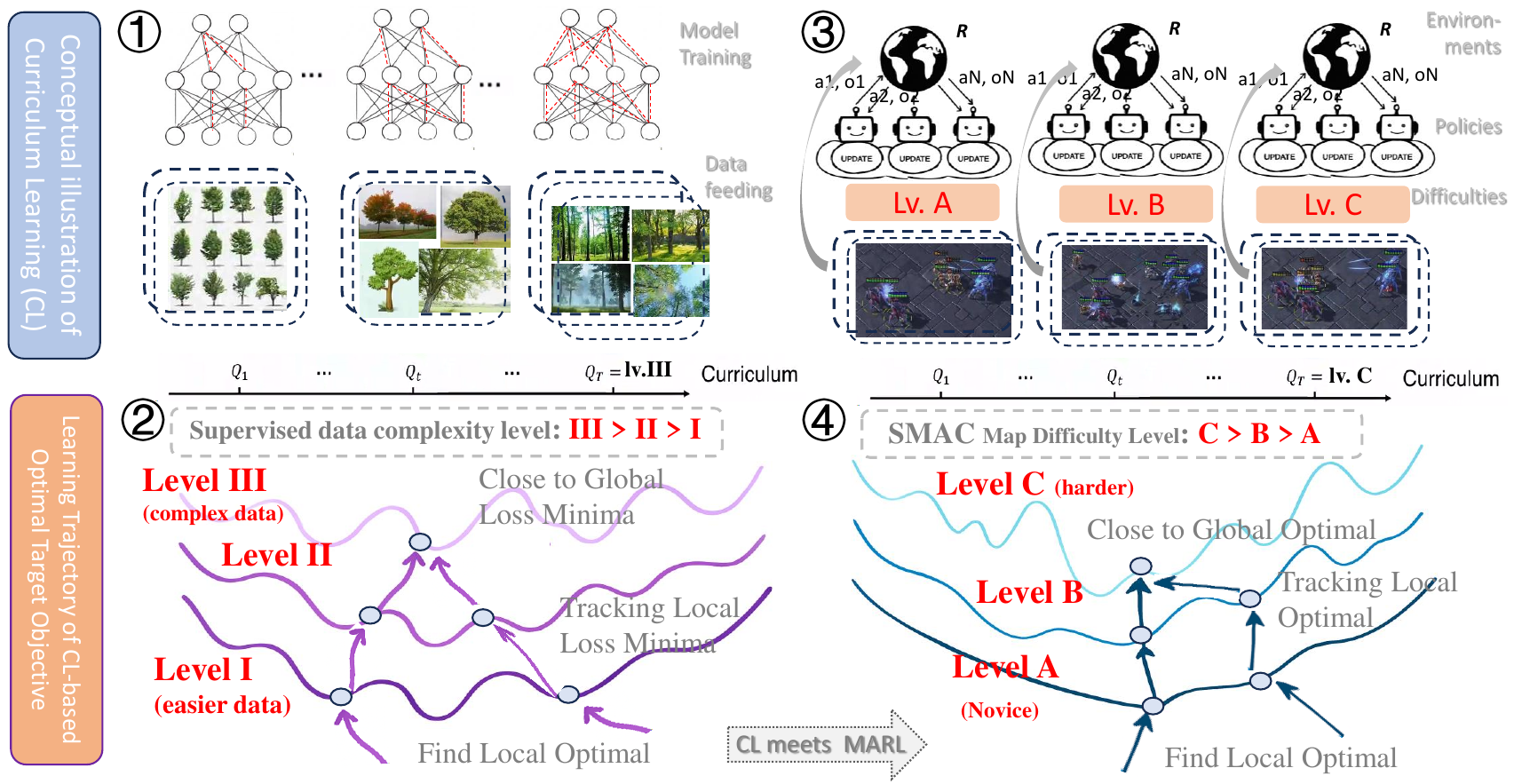}
\caption{\textit{When CL Meets MARL}: A comparative illustration of training paradigms between CL-based supervised learning and CL-based MARL. Part 1 and Part 2 represent the traditional CL framework for supervised learning (e.g., CV, NLP) with labeled data and its optimization trajectories (supervision loss minimization), and Part 3 and Part 4 are the CL-based MARL training framework in the SMAC adversarial environment (\texttt{2s\_vs\_3z} map) and corresponding optimization trajectories (optimal policy exploration).} 
\label{fig_CLmeetsMARL}
\end{figure*}

Previous studies have continuously addressed several existing key challenges in MARL, such as sparse global rewards, inefficient credit assignment among agents and insufficient environmental exploration \cite{WeightedQMIX,EMCzheng2021episodic,emu2024}, leading to substantial progress in algorithm design and performance.
Crucially, the limitations targeted by these efforts all live at the level of \textit{within-run dynamics}, whereas an orthogonal limitation persists at the \textit{task-distribution level}: the overall task difficulty itself is held fixed across an entire training run.
Beyond these well-studied issues, this orthogonal limitation has been largely overlooked: most MARL models are trained under meta-stationary environmental conditions, which restrict policy adaptability. As highlighted in recent reviews \cite{SindhuDRLdynENVs,hernandezleal2019survey}, widely adopted MARL benchmarks often assume fixed opponent behavior\footnote{We use \textit{environmental meta-stationarity of MARL} to denote a property of the \textit{task distribution}: the overall difficulty level of the task (e.g., opponent AI strength, traffic flow complexity) is held fixed across the entire training run. This is conceptually \textit{orthogonal} to the \textit{non-stationarity} discussed in MARL theory \cite{hernandezleal2019survey}, which refers to shifts in transition or reward dynamics \textit{within} a single training run, induced by simultaneously learning agents. The two coexist on different axes: a non-stationary learning process can still operate under meta-stationary task conditions, and meta-stationarity does not eliminate within-run non-stationarity. For instance, in military simulation the opponent AI capability is fixed (meta-stationary) while agents' co-adapting policies still make the joint dynamics non-stationary; in traffic control, the overall traffic flow complexity stays constant across episodes while local conditions still fluctuate.}.
For instance, in the challenging StarCraft-II multi-agent environment of the \textit{StarCraft multi-agent challenge} (SMAC) benchmark\footnote{SMAC GitHub: \url{https://github.com/oxwhirl/smacv2}.},
the game engine configures scripted opponents with predefined difficulty levels.
Lower levels (e.g., Levels 1–3) serve as the `Easy' modes, while higher levels (e.g., Levels 8–10)
employ `Cheating' modes with access to global vision, faster reactions, and additional resources.
However, the SMAC environment \cite{ellis2023smacv2} through its game engine API, exclusively defaults to Level 7 difficulty across all SMAC map scenarios \cite{WeightedQMIX,EMCzheng2021episodic,emu2024,MARR}. 
Throughout this work the opponents are \textit{scripted}, not learning: they execute SMAC's built-in heuristic AI at the discrete difficulty level set by FlexDiff, in contrast to \textit{self-play} settings (e.g., AlphaStar) where the adversary is itself a learning agent.

While such meta-stationarity environments help maintain training stability and consistent policy convergence during training, 
their reliance on fixed opponent capabilities (e.g., the default Level-7 difficulty) poses a critical limitation: agents often overfit to static conditions and fail to generalize across dynamic scenarios.
Consequently, agents frequently converge to suboptimal local solutions that perform well under static conditions but fail to adapt beyond them.
Unlike the single-agent setting, MARL introduces additional challenges including the environmental dynamics, exponential joint action space growth, \textit{credit assignment rationality}, and \textit{partial observability}, all of which are \textit{aggravated rather than resolved} when training is also meta-stationary, since a fixed task distribution prevents agents from encountering the variation needed to develop transferable coordination. 
These complexities are further exacerbated under fixed environmental dynamics.
This observation motivates us to consider a fundamental question:
\begin{hypobox}
\small
``\textit{Can we design a human-inspired progressive learning paradigm for the cooperative adversarial tasks of MARL that mimics the developmental stages of human skill acquisition to overcome the `environmental meta-stationarity' limitation?}''
\end{hypobox}
In recent years, curriculum learning (CL) has gained significant attention as a progressive learning paradigm that mimics human learning process by moving from easier to more difficult tasks, 
particularly succeeds in supervised learning, e.g., computer vision (CV) and natural language processing (NLP) \cite{CurriculumSurvey}. As demonstrated in the Part 1 of Fig. \ref{fig_CLmeetsMARL}, compared with direct training on the entire dataset, CL strategically partitions the training data into difficulty-graded subsets, progressively increasing challenge levels during training, mirroring human educational curricula. 
From an optimization perspective shown in the Part 2 of Fig. \ref{fig_CLmeetsMARL}, 
the training process first converges to a local minimum on a simpler and easier subset, and as the curriculum progresses, it continues to follow these minima while adapting to more complex objectives.
This mechanism ensures sustained generalization improvement, ultimately enabling the model to approximate the global optimum.

Inspired by these advantages, we systematically investigate CL's potential for optimizing multi-agent cooperative adversarial MARL training. To the best of our knowledge, no research has been conducted on implementing CL for MARL's cooperative adversarial tasks such as SMAC before this work. Our research effectively mitigates the long-overlooked limitations caused by the `\textit{environmental meta-stationarity of MARL}' phenomenon, poor generalization and local optima convergence pitfalls. Specifically, when applying CL to MARL, we identify key parallels with traditional supervised CL approaches. As illustrated in the Part 3 of Fig. \ref{fig_CLmeetsMARL}, progressively increasing environmental difficulty for multi-agent systems enables policy learning to follow the same easy-to-hard progression. From a MARL policy optimization perspective (Part 4 of Fig. \ref{fig_CLmeetsMARL}), agents first converge to local optima in simpler and easier environments.
As the task difficulty escalates, the policies continuously adapt by tracking solution manifolds across progressively challenging environments, ultimately approaching the global optimum in the target difficulty environment.  

To make CL effective in the MARL context, two fundamental components must be addressed, i.e., the design of a \textit{difficulty measurer} and a \textit{training scheduler} \cite{CurriculumSurvey}.
The fundamental challenges in CL lie in the implementation of the difficulty Measurer and the determinization of the training scheduler \cite{CurriculumSurvey}\footnote{In CL-based supervised learning (e.g., CV and NLP). The difficulty measurer focuses on partitioning the dataset into different difficulty levels and classifying each supervised sample. And the training scheduler determines when and how to introduce harder samples.}. 
In CL-based MARL, the difficulty measurer defines how to modify the task complexity, which can be achieved by tuning environment or task difficulty through hyperparameter configuration
(e.g., varying opponent AI levels in SMAC's StarCraft II scripting interface). The MARL \textit{training scheduler} determines when to adjust overall difficulty, where evident signals like overall rewards and task completion rates (e.g., SMAC rewards and win rates) serve as criteria for comprehensive task difficulty adjustment.

To implement this, we design a statistical-based adaptive training scheduler, namely `\textit{flexible task difficulty scheduler}' (\textbf{FlexDiff}), that dynamically modulates MARL task difficulty to optimize policy learning. 
Specifically, \textbf{FlexDiff} dynamically adjusts environmental task difficulty based on real-time agent performance metrics, particularly the rewards and task completion rates.
The proposed module integrates three components, i.e., (1) momentum-driven adjustment, capturing persistent performance trends, (2) sliding window analysis, evaluating short-term stability via mean and std metrics, and (3) dual-curve monitoring, simultaneously tracking reward progression slopes and task completion rate derivatives. 
By synthesizing these multi-scale features through adaptive thresholding, the scheduler self-regulates difficulty levels in real time, achieving flexible curriculum adaptation to emergent agent capabilities while maintaining training stability against local optima\footnote{We have thoroughly investigated the design of MARL training scheduler. Given that the data volume of task completion and cumulative rewards during early or even entire MARL training stages is typically insufficient to train converged and stable ML-based (including RL-based) Difficulty Schedulers. 
Furthermore, the mutual dynamic influence between MARL and the training scheduler's learning process may lead to unstable training conditions.
Compared to these ML-based strategies, statistical methods with parameters like window size and thresholds possess clear physical meanings and offer more interpretable logic than machine learning black-box models. More importantly, statistical-based training scheduler, such as linear regression, features low computational complexity, enabling stable and effective environment difficulty adjustment without additional training overhead.}. 

While CL improves MARL agents' generalization, merely increasing difficulty via training feedback is suboptimal due to (1) post-transition reward sparsity and (2) policy rigidity. Agents converging to Difficulty I often fail to adapt to Difficulty II. We address this by prioritizing high-contributing agents via optimized credit assignment.
Specifically, inspired by \textit{group relative policy optimization} (GRPO) \cite{shao2024deepseek} and \textit{counterfactual multi-agent policy gradients} (COMA) \cite{FoersterCOMA} algorithms, we propose a novel \textit{Counterfactual Group Relative Policy Advantage} (\textbf{CGRPA}) algorithm that evaluates intra-agent action advantages by quantifying each agent's environmental contribution through comparison between actually executed actions from empirical trajectories, and counterfactual actions that were possible but not actually selected. This algorithm helps identify key contributors at each timestep, thereby stabilizing convergence during difficulty transitions.
\textbf{CGRPA} uniquely combines GRPO's group optimization algorithm with COMA's counterfactual theory, effectively addressing the non-convergence issues and significantly enhancing the agents' policy optimization speeds. 
To our knowledge, this is the first technical integration of GRPO with COMA's counterfactuals in MARL.
Through extensive experiments on SMAC, we demonstrate that \textbf{CGRPA} effectively mitigates policy divergence risks during environmental difficulty transitions and enhances the coordinated implementation of CL in MARL.

In summary, our work builds directly off of the idea of CL paradigm and reward credit assignment enhancements.
Through rational technological innovations, we have effectively overcome the widespread challenge of environmental meta-stationarity in MARL.
The primary contributions of this paper are summarized as follows:

\begin{itemize}
\item 
Through systematic analysis of existing MARL research, we firstly identify the prevalent challenge of `environmental meta-stationarity' in MARL, which fundamentally limits agents' policy generalization and constrains agents' ability to discover globally optimal joint policies.
\item 
To address this issue, inspired by CL from supervised learning, we propose a novel adaptive \textit{training scheduler} for MARL, namely \textbf{FlexDiff}, that dynamically adjusts environmental task difficulty during training.
To the best of our knowledge, this represents the first integration of CL training paradigm into the multi-agent cooperative adversarial scenarios of MARL, particularly in the testbed of StarCraft-II unit micromanagement, SMAC.
\item 
To mitigate the non-convergence risks of MARL introduced by CL integration, we propose the \textbf{CGRPA} algorithm, a novel approach that combines GRPO's group optimization with COMA's counterfactual reasoning. By evaluating intra-agent advantages through the comparison between empirical and counterfactual actions, \textbf{CGRPA} stabilizes policy adaptation during difficulty transitions. 
\item
Through extensive empirical evaluation across multiple SOTA MARL baselines, such as QMIX \cite{QMIX}, OW-QMIX \cite{WeightedQMIX}, DER \cite{DERhu2023discriminative}, and EMC \cite{EMCzheng2021episodic}, our CL-enhanced MARL approach demonstrates significant performance improvements, achieving competitive results compared to SOTA algorithms. For instance, experiments on SMAC benchmark show that QMIX's win rate remarkably surges from 45-60\% to 75-90\% in \texttt{5m\_vs\_6m} map.
These results validate the effectiveness of CL-MARL and reveal CL's significant potential for MARL research.  
\end{itemize}

\section{Progress in MARL} 
MARL, as an extension of RL under a single-agent context to more complex cooperative scenarios, has consistently garnered significant attention. Compared to single-agent RL, MARL inevitably faces several issues, 
such as: 
\textit{Non-stationarity in Global States} \cite{li2022dealing,weiTackling,pmlrv232nekoei23a,sun2022monotonic}, 
\textit{Credit Assignment Problems} \cite{FoersterCOMA,QMIX,QTRAN,VDN}, \textit{Partial Observability} \cite{Guocommunication}, and \textit{High-Dimensional Action Space} \cite{zhang2023addressing,MINGACTION2023281}. 
Additionally, other critical challenges in MARL, such as \textit{Exploration-Exploitation Trade-off} \cite{EMCzheng2021episodic}, and \textit{Scalability and Generalization} \cite{MARR}, have been continuously researched as they play an essential role in the effective MARL optimization. 
Next, we will provide a brief overview of recent advancements in addressing these challenges. 

For dealing with non-stationarity of global states, Nekoei et al. \cite{pmlrv232nekoei23a} proposed a multi-timescale learning approach in decentralized cooperative MARL, where agents update their policies at different rates, improving convergence speed and minimizing non-stationarity. Wei et al. \cite{weiTackling} formalized the non-stationary environment using pessimistic transitions and introduced prudent Q-learning (PQL) to control multi-agent systems independently in such environments. Sun et al. \cite{sun2022monotonic} introduced a monotonic improvement guarantee for MARL by enforcing trust region constraints over joint policies, providing a theoretical foundation for proximal ratio clipping.

To better address the \textit{Credit Assignment Problems} in the \textit{Centralized Training with Decentralized Execution} (\textbf{CTDE}) paradigm, QMIX \cite{QMIX} optimizes VDN \cite{VDN} by introducing a hypernetwork that combines local value functions with global state information, replacing the simple summation of local Q-values in VDN. 
Foerster et al. \cite{FoersterCOMA} proposed COMA, which employs a centralized critic with a counterfactual advantage function that marginalizes each agent's action while holding others fixed, enabling precise credit assignment in MARL.

For the efficient experience utilization, 
Yang et al. \cite{yang2024sampleefficient} proposed MARR, a novel framework that enhances experience utilization efficiency in multiagent systems through a reset strategy and data augmentation technique, significantly improving sample efficiency in MARL.
Moreover, several studies enhanced learning experience efficiency in RL by incorporating episodic memory buffers \cite{EMCzheng2021episodic,emu2024}. These approaches record optimal state-action Q-values from historical trajectories, serving as retrievable memory signals to enhance learning from experience buffers.

In addition to the aforementioned approaches, recent works tackled existing problems such as partial observability and complex environmental dynamics through enhanced inter-agent communication mechanisms \cite{Guocommunication} that facilitate efficient information sharing, as well as developing innovative agent grouping strategies \cite{ShaogroupMARL}
employing hierarchical architectures to effectively integrate localized observations. 

In cooperative-adversarial MARL scenarios, the most widely adopted benchmark environments include SMAC \cite{ellis2023smacv2}, google research football (GRF) \cite{Kurach2020gfootball}, and Neural MMO \cite{suarez2021neuralmmo}. 
Empirical reviews indicated that approximately 72\% of recent MARL publications utilize SMAC, while nearly 58\% employ GRF as primary testbeds \cite{papoudakis2021benchmarking,MARL2025surveymultiagent}. 
Among them, SMAC \cite{ellis2023smacv2}, based on \textit{StarCraft-II, Blizzard Entertainment}\texttrademark, has become the benchmark environment for evaluating multi-agent cooperative-adversarial algorithms due to its 40+ heterogeneous battle scenarios and flexible adversarial mechanics. By featuring diverse unit compositions and adjustable AI difficulty levels, it provides an ideal testing platform for MARL policy development.

\paragraph*{Curriculum learning in MARL}
A small but growing line of work has applied CL to (multi-agent) RL, and our setting is best understood by contrast with these approaches. Narvekar et al.\ \cite{narvekar2020curricSurvey} provide the canonical framework for CL in RL, formalising a curriculum as a sequence of source tasks selected via an explicit task graph or curriculum-MDP; Portelas et al.\ \cite{portelas2020ACLsurvey} survey automatic curriculum learning (ACL) methods that adapt task distributions to agent capability. Within MARL, two representative population-based curricula are DyMA-CL by Wang et al.\ \cite{wangAAAI2020dyma}, which trains agents on small teams and progressively increases team size on SMAC, and EPC by Long et al.\ \cite{longICLR2020EPC}, which evolves agent populations across stages of growing scale on cooperative-competitive tasks. A third common pattern is self-play curricula, exemplified by AlphaStar's league training \cite{vinyals2019alphastar}, where the opponent is bootstrapped from past versions of the agent itself. CL-MARL departs from each of these along complementary axes. Unlike DyMA-CL and EPC, which alter the agent population (team size or composition), CL-MARL keeps the team fixed and instead modulates the \textit{built-in opponent strength}, leaving the policy and value networks architecturally untouched. Unlike the curriculum-MDP framing of Narvekar et al., FlexDiff requires no learned task selector or hand-built task graph: the difficulty index is moved by closed-form statistics computed directly from the win-rate and reward streams. Unlike self-play curricula, the scripted SMACv2 AI provides a stationary opponent family at each level, so curriculum stages are reproducible across runs and seeds. The combination of (i) opponent-strength modulation, (ii) a statistics-driven scheduler with no learned components, and (iii) a tightly coupled credit-assignment companion (CGRPA) for transition stability is, to our knowledge, not present in prior CL-for-MARL work.

Despite significant progress in MARL achieved through both algorithmic innovations and benchmark development \cite{Xu9904958SurveyDRL,NING2024Survey,MARL2025surveymultiagent}, a fundamental limitation still persists: ``\textit{environmental meta-stationarity}''. In other words, most current MARL methods are evaluated under fixed-difficulty adversarial conditions (e.g., SMAC's default Level-7 AI opponents). Such meta-stationary conditions often leads agents to develop environment-specific strategies that may result in environment-overfitted policies with poor generalization and convergence to locally optimal but globally suboptimal strategies.
Hence, the primary objective of this work is to systematically integrate CL \cite{CurriculumSurvey} into cooperative-adversarial MARL, addressing the long-overlooked bottleneck of ``\textit{environmental meta-stationarity}''. 

\section{Preliminaries}
\subsection{Multi-Agent Reinforcement Learning}
MARL is essentially a sequential decision-making task under multi-agent cooperative context, which can be formally modeled as a \textit{decentralized partially observable markov decision process} (\textbf{Dec-POMDP}) \cite{Oliehoek2016POMDP}. 
Formally, a Dec-POMDP is defined by the tuple $G = \langle \mathcal{S}, \mathcal{U}, P, r, \Omega, O, n, \gamma \rangle$, where $\mathcal{S}$ is the state space, $\mathcal{U}$ is the action space for each of the $n$ agents, $P(s'|s,\boldsymbol{u}): \mathcal{S} \times \mathcal{U}^n \times \mathcal{S} \to [0,1]$ is the state transition function, $r(s,\boldsymbol{u}): \mathcal{S} \times \mathcal{U}^n \to \mathbb{R}$ is the shared reward function, $\Omega$ is the observation space with emission probabilities $O(s,a): \mathcal{S} \times \mathcal{A} \to \Omega$, and $\gamma \in [0,1)$ is the discount factor.
At each timestep $t$, each agent $a \in \{1,...,n\}$ selects an action $u^a \in \mathcal{U}$, forming a joint action $\boldsymbol{u} \in \mathcal{U}^n$. The environment transitions according to $P(s_{t+1}|s_t,\boldsymbol{u}_t)$, generating a scalar reward $r_t = r(s_t,\boldsymbol{u}_t)$. Agents receive individual observations $z^a_t \sim O(s_t,a)$ that may represent partial state information.

Each agent maintains an action-observation history $\tau^a_t \in \mathcal{T} = (\Omega \times \mathcal{U})^*$, which maps to actions through a stochastic policy $\pi^a(u^a|\tau^a): \mathcal{T} \times \mathcal{U} \to [0,1]$. The joint policy $\boldsymbol{\pi}$ induces a centralized action-value function:
\begin{equation}
    Q^{\boldsymbol{\pi}}(s_t,\boldsymbol{u}_t) = \mathbb{E}_{s_{t+1:\infty},\boldsymbol{u}_{t+1:\infty}} \left[ \sum_{i=0}^\infty \gamma^i r_{t+i} \Big| s_t, \boldsymbol{u}_t \right].
\end{equation}

While training can leverage centralized information (full state $s$ and all agents' histories $\boldsymbol{\tau}$), execution must be decentralized - each agent's policy $\pi^a$ depends only on its local history $\tau^a$. This framework subsumes both the fully observable MMDP case (when $O(s,a) = s$) and standard POMDPs (when $n=1$). The key challenge emerges from the exponential growth of joint action space $\mathcal{U}^n$ and the partial observability constraints during execution.

MARL algorithms are typically categorized into three training paradigms: \textit{centralized training and execution} (\textbf{CTCE}) \cite{DQNori,DQNv3}, \textit{decentralized training and execution} (\textbf{DTDE}) \cite{claus1998IQL}, and \textit{centralized training with decentralized execution} (\textbf{CTDE}) \cite{amato2024CTDEsurvey}. 
Among these, CTDE has become predominant due to its superior learning efficiency while maintaining the flexibility of decentralized execution. 
The algorithms under CTDE paradigm combine decentralized execution where each agent $a$ acts via its own policy $\pi^a(u^a|\tau^a)$ conditioned on local action-observation histories $\tau^a = (z_0^a,u_0^a,...,z_t^a)$, with centralized training that optimizes a joint action-value function $Q_{tot}(\tau,u,s) = g(\{Q^a(\tau^a,u^a)\}_{a=1}^n, h(s))$ where: (1) individual utilities $Q^a$ depend only on local $\tau^a$, (2) the state encoder $h(s)$ integrates global information, and (3) the mixing network $g(\cdot)$ enforces consistency between decentralized policies and centralized value estimation through the monotonicity constraint $\partial Q_{tot}/\partial Q^a \geq 0$.
Notable algorithms including VDN \cite{VDN}, QMIX \cite{QMIX}, MAPPO \cite{MAPPO}, and MADDPG \cite{MADDPG} all adhere to CTDE. Similarly, our proposed method is also based on the baseline algorithms of CTDE. 
Throughout this paper, the underlying optimisation objective remains the discounted cumulative reward defined in Eq.~(1); the win rate $w_i$ that appears later in §IV-A is consumed only as an external evaluation and curriculum-scheduling signal by FlexDiff, not as an additional term in the policy-gradient or value-update objective. For benchmarks lacking a binary success indicator, the win-rate channel can be omitted and FlexDiff degenerates to its reward-only branch (Eqs.~\eqref{nabla2r}--\eqref{mtupdate}) without altering Eq.~(1).

\subsection{Curriculum Learning (CL)}
\textbf{CL} \cite{CurriculumSurvey} is a novel continuous learning paradigm originally proposed by Bengio et al. \cite{BengioCLori} for supervised learning tasks in CV and NLP. 
This paradigm progressively exposes a ML model to data samples of increasing complexity, mimicking the structured learning process in human education. 
Formally, in supervised training, given a training dataset $\mathcal{D} = \{(x_i, y_i)\}_{i=1}^{\lVert \mathcal{D} \rVert}$ where $x_i \in \mathcal{X}$ is an input (e.g., an image or text sequence), $y_i \in \mathcal{Y}$ is its corresponding label, and $\lVert \mathcal{D} \rVert$ denotes the sample size, CL introduces a \textit{difficulty measure} $d: \mathcal{X} \times \mathcal{Y} \rightarrow \mathbb{R}^+$ that quantifies the complexity of each training.

The learning process is governed by a \textit{curriculum scheduler},or \textit{training scheduler}, $\mathcal{C}(t)$, which defines a sequence of training distributions $\{Q_t\}_{t=1}^T$ over $\mathcal{D}$, let $T$ denote the number of curriculum stages. At each training stage $t$, the model optimizes a reweighted objective:
\(
\min_\theta \mathbb{E}_{(x,y) \sim Q_t} \left[ \mathcal{L}(f_\theta(x), y) \right],
\) where $f_\theta$ is the model with parameters $\theta$, $\mathcal{L}$ is the loss function, and $Q_t$ is a distribution that prioritizes easier examples early in training before gradually incorporating harder ones. The scheduler ensures that the effective training data transitions smoothly from a simpler subset $\mathcal{D}_1 \subset \mathcal{D}$ to the full dataset $\mathcal{D}_T = \mathcal{D}$, satisfying:
\begin{equation}
\mathbb{E}_{Q_t}[d(x,y)] \leq \mathbb{E}_{Q_{t+1}}[d(x,y)], \quad \forall t \in \{1, \dots, T-1\}.
\end{equation}

Formally, the implementation of \textit{curriculum scheduler} can be defined $Q_t$ as a reweighting of the original data distribution $P(x,y)$:
\(
Q_t(x,y) = w_t(x,y) \cdot P(x,y),
\)
where $w_t(x,y)$ is a time-dependent weight function that increases for harder examples as $t$ progresses.
For instance, in \textit{self-paced learning} \cite{NIPS2010SelfPaced}, a CL variant, its weights are binary ($w_t \in \{0,1\}$) and determined by a threshold $\tau_t$ on difficulty: 
\(
w_t(x,y) = \mathbb{I}[d(x,y) \leq \tau_t], \quad \text{with} \quad \tau_t \leq \tau_{t+1}.
\)
Noteworthy, the final effectiveness of curriculum is determined by the alignment between $d(x,y)$ and the model's current capability, examples deemed `\textit{easy}' should be those the model can learn reliably at stage $t$, while `\textit{hard}' examples require more mature representations. In NLP, $d(x,y)$ might measure sentence length or syntactic complexity. In CV, it could reflect image clutter, label noise, or distributional mismatch. The scheduler $\mathcal{C}(t)$ may be predefined (e.g., linear annealing of $\tau_t$) or adaptively tuned based on model performance. 

\section{Progressive CL for MARL}
\begin{figure*}[ht]
\centering
\includegraphics[width=1.0\linewidth]{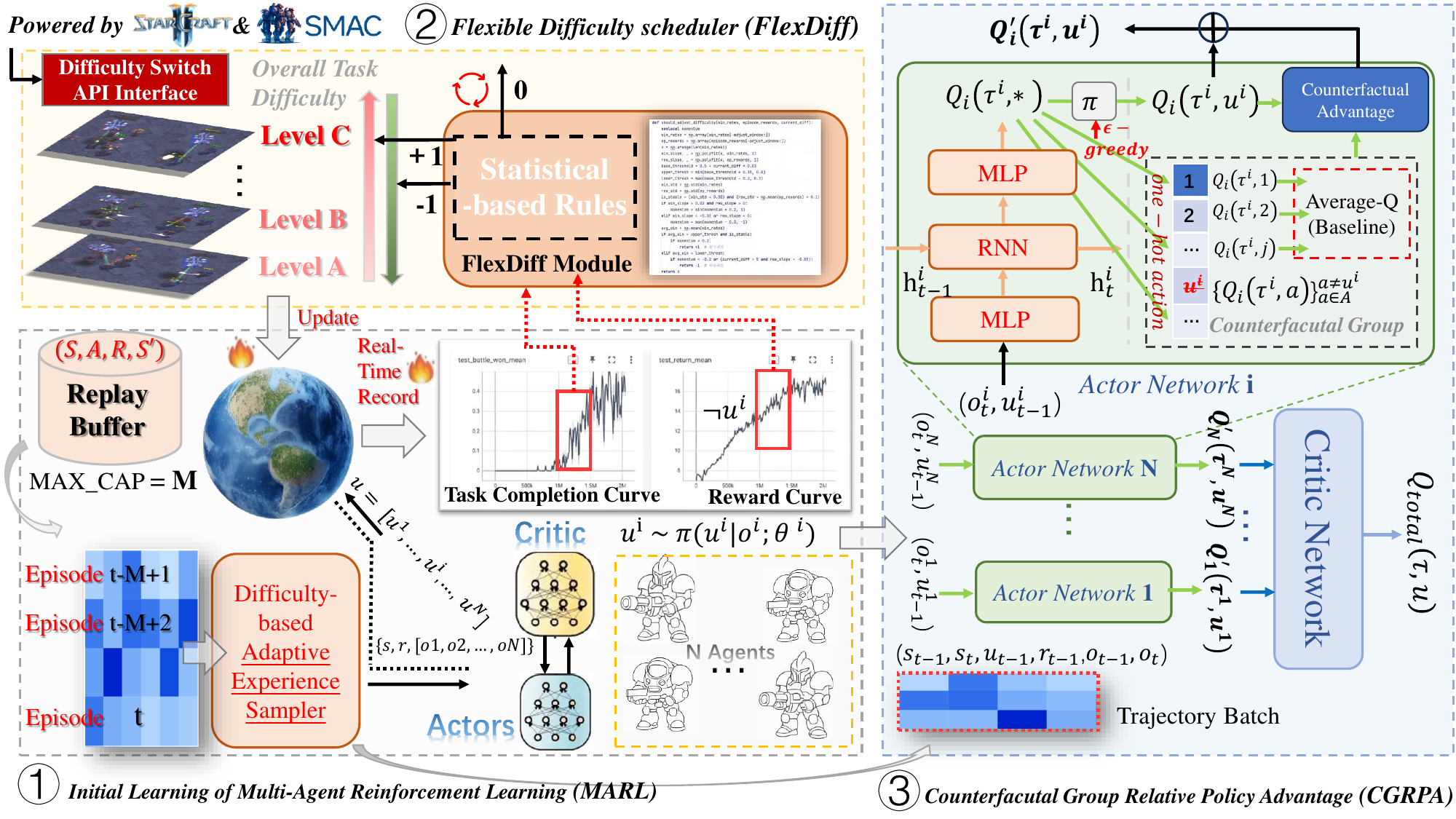}
\caption{The architecture of our proposed CL-based MARL framework for cooperative-adversarial scenarios in SMAC. The framework comprises: (1) the original MARL training with static environment settings, (2) our proposed \textbf{FlexDiff} scheduler for dynamic difficulty adjustment, and (3) our introduced CGRPA algorithm enhancing \textit{Actor-Critic} with counterfactual inter-action advantage estimation.}  
\label{fig_ourmodel}
\end{figure*}

In this section, we propose an innovative algorithmic innovation designed to enhance MARL baseline methods through CL. 
We refer to this approach as the \textit{CL-enhanced MARL framework} (CL-MARL).
As illustrated in Fig. \ref{fig_ourmodel}, our approach integrates two novel components, i.e., \textbf{FlexDiff} and \textbf{CGRPA}.
The framework addresses two critical challenges, i.e., the inherent limitations of ``\textit{environmental meta-stationarity}'' in conventional MARL paradigms, and the \textit{training instability} caused by abrupt environmental difficulty transitions.

\subsection{Flexible Difficulty Scheduler}  
\label{FlexDiffsec}
The fundamental challenges in CL lie in the implementation of the \textit{difficulty measurer} and the determinization of the \textit{training scheduler}. 

\subsubsection{The Implementation of Difficulty Measurer for SMAC}
In CL for supervised tasks such as CV and NLP, the difficulty measurer partitions data into levels based on sample complexity \cite{CurriculumSurvey}.
For CL-based MARL, difficulty is defined through task-level variations, commonly by adjusting environment parameters. For SMAC, this involves modifying the opponent AI difficulty via the StarCraft II scripting interface.


Rather than requiring pre-configuration within the StarCraft II engine, the \textit{difficulty measurer} in SMAC leverages a built-in API to directly adjust opponent AI levels during training. This enables flexible and automated difficulty modulation without manual scripting. As illustrated in Part 2 of Fig. \ref{fig_ourmodel}, the API provides 10 discrete difficulty levels.
The built-in AI opponents difficulty levels are: \textit{1 (Very Easy), 2 (Easy), 3 (Medium), 4 (Hard), 5 (Harder), 6 (Very Hard)}, and \textit{7 (Elite)}. Beyond these, levels 8–10 are ``cheat-level'' AIs that gain unfair advantages: \textit{8 (Map hack)}, \textit{9 (Resource Cheating)}, and \textit{10 (Map hack + Resource Cheating)}, where level 10 approximates the skill of top 30–50\% human players \cite{TENCENTsun2018tstarbots}.
This API-based design makes the difficulty measurer both efficient and adaptable for curriculum-based MARL training.
For details, refer to Blizzard’s API documentation or in-game AI settings\footnote{Blizzard’s Documentation: \url{https://develop.battle.net/documentation}.}. 

\subsubsection{The Implementation of \textbf{FlexDiff}} 
Typically, the training scheduler under supervised learning setting determines when and how to introduce harder samples for training \cite{CurriculumSurvey}. 
Correspondingly, the training scheduler in MARL context depends when to increase or decrease overall environmental difficulty, i.e., opponents' strength in multi-agent adversarial scenarios, where evident signals during agents training serve as criteria for task difficulty adjustment.

Hence, our proposed adaptive training difficulty scheduler, i.e., \textbf{FlexDiff}, dynamically modulates environmental difficulty based on real-time agent performance, primarily relying on the real-time obtained rewards and task completion rates (e.g., SMAC average rewards and win rates).

As illustrated in Part 2 of Fig. \ref{fig_ourmodel}, \textbf{FlexDiff} achieves self-adaptive CL through analyzing multi-scale statistical feature, with three key innovations: \textit{synergistic dual-metric evaluation, momentum-driven dynamic adjustment buffer}, and \textit{dynamic boundary constraints} for training stability.
We formalize its core mechanisms as follows:

\textbf{Synergistic dual-metric performance evaluation:}
A single performance scalar suffers from high estimator variance under stochastic policy rollouts, and variance-controlled estimators are known to stabilise stochastic-approximation updates~\cite{borkar1997twotimescale,polyak1992ruppert}. We therefore evaluate two statistically complementary signals: a binary success indicator (win rate $w_i$) and a continuous return ($r_i$). The win rate is a Bernoulli statistic with bounded variance and direct task-completion semantics, while the return captures graded progress that the binary signal discards once policies approach saturation. Estimating the mean and variance of both metrics jointly yields a strictly more informative test of curriculum readiness than either signal alone, because advancing a curriculum stage requires high mean (competence) together with low variance (reliability). Each evaluation runs $K{=}32$ rollouts at the active difficulty, and the resulting pairs $(w_i,r_i)$ are appended to a fixed-length history that feeds the sliding-window analysis below.

To monitor the agents' recent training progress, we employ two fixed-length sliding windows to track the performance metrics of win rates and averaged rewards. Formally, we define the performance evaluation window $\mathcal{W}_t$ at training step $t$ as:
\begin{equation}
\mathcal{W}_{t} = \{ (w_i, r_i) \}_{i=t-N+1}^t,
\end{equation}
where $w_i \in [0,1]$ represents the normalized win rate (victories divided by total testing episodes) at the $i_{\rm th}$ evaluation cycle, and $r_i \in \mathbb{R}^+$ denotes the correspondingly episode reward.

The overall stability of agents' performance is formalized through the following constraints: 
\begin{align}
\label{sigmaw2}
\sigma_w^2 &= \frac{1}{N}\sum_{i=1}^N (w_i - \mu_w)^2,\quad \mu_w = \frac{1}{N}\sum_{i=1}^N w_i,
\end{align}
\begin{align}
\sigma_r^2 &= \frac{1}{N}\sum_{i=1}^N (r_i - \mu_r)^2,\quad \mu_r = \frac{1}{N}\sum_{i=1}^N r_i, \\
\mathcal{S} &= \mathbb{I}(\sigma_w < \text{L}_{toler}) \otimes \mathbb{I}(\sigma_r < \text{R}_{toler}), 
\end{align}
where $\mathbb{I}$ acts as a binary classifier for stability conditions, $A\otimes B$ denotes the simultaneous satisfaction of both constraints $A$ and $B$. The stability indicator $\mathcal{S} \in \{0,1\}$ combines low value in both success rates and rewards, ensuring reliable policy evaluation before difficulty adjustments.

For the windowed win rates ${w_i}$, $i=1,\ldots,N$, the temporal trend $\beta_w$ is quantified through linear regression:
\begin{equation}
\label{betaw}
\beta_w = \frac{N\sum x_iw_i - \sum x_i \sum w_i}{N\sum x_i^2 - (\sum x_i)^2} \,v.s.\, \zeta,\; \; x_i = i - t + N - 1,
\end{equation}

In Eq.~\eqref{betaw}, the slope $\beta_w$ separates three operational regimes: improvement ($\beta_w>\zeta$), stationarity ($|\beta_w|\le\zeta$), and degradation ($\beta_w<-\zeta$). The \textit{momentum threshold} $\zeta$ is defined as the smallest slope statistically distinguishable from sampling noise; its concrete numerical value, recommended range, and sensitivity analysis are deferred to Sec.~\ref{settings} to keep the methodological discussion uncluttered. The summation index $\sum$ runs over the $N$ window samples.
Then, we compute second-order differences for reward to detect phase transitions:
\begin{equation}
\label{nabla2r}
\nabla^2 r_t = r_t - 2 \cdot r_{t-1} + r_{t-2}.
\end{equation}

In Eq. \eqref{nabla2r}, positive values ($\nabla^2 r_t > 0$) indicate convex acceleration where marginal improvements grow exponentially, characteristic of effective policy refinement. 
Near-zero values ($\nabla^2 r_t \approx 0$) indicate linear progress with a constant learning rate, while negative values ($\nabla^2 r_t < 0$) signal a slowdown in convergence, where returns decrease, often preceding a performance collapse.

\textbf{Momentum-driven dynamic adjustment buffer:}
The two trend signals $\beta_w$ and $\nabla^2 r_t$ are individually noisy and would induce chattering if used as raw decision variables. We filter them through an exponential moving average (EMA), the same low-pass filter used in Adam~\cite{kingma2015adam} and in Polyak--Ruppert averaging~\cite{polyak1992ruppert}, which provably reduces the variance of stochastic-approximation iterates. The resulting momentum term $m_t\!\in\![-1,1]$ is updated as
\begin{equation}
\label{mtupdate}
m_t = \gamma_m\, m_{t-1} + (1-\gamma_m)\,\tanh\!\big(\beta_w + \tfrac{1}{2}\nabla^2 r_t\big),
\end{equation}
where $\gamma_m\!\in\![0,1)$ is the EMA decay coefficient with effective memory horizon $\approx 1/(1-\gamma_m)$ evaluation cycles. We use $\gamma_m$ rather than $\gamma$ to disambiguate this coefficient from the discount factor $\gamma$ defined in Sec.~\ref{sec:cgrpa}. The $\tanh(\cdot)$ wrapper bounds each instantaneous update to $[-1,1]$, a standard saturation device in robust control that prevents an outlier window from dominating $m_t$. The combined trend signal $\beta_w + \tfrac{1}{2}\nabla^2 r_t$ fuses first- and second-order information so that $m_t$ rises only when both win rate and reward improve. The concrete value of $\gamma_m$ together with its sensitivity is reported in Sec.~\ref{settings}.

\textbf{Dynamic boundary constraints:}
Promotion and demotion errors are asymmetric in cost. A premature promotion exposes a half-trained policy to an opponent it cannot match, which can trigger \textit{catastrophic policy unlearning}: the value network bootstraps from rewards generated by what is effectively a random policy, and previously learned behaviour is overwritten before recovery is possible. A premature demotion only slows the curriculum by a few cycles. The expected regret of the two errors therefore differs by orders of magnitude, and this regret asymmetry justifies an asymmetric decision boundary. We instantiate the principle through difficulty-adaptive thresholds:
\begin{align}
\tau_h(d_t) &= \min(\varphi_{\max}, \text{Diff}_{anch} + \text{Scal}_{coef} \cdot d_t), \\
\tau_l(d_t) &= \max(\varphi_{\min}, \text{Diff}_{anch} - \text{Scal}_{coef} \cdot d_t).
\end{align}

Here $\text{Diff}_{anch}$ is the anchor reference performance level and $\text{Scal}_{coef}$ is the per-step scaling. The asymmetry is enforced by capping the promotion threshold above ($\varphi_{\max}$) and the demotion threshold below ($\varphi_{\min}$): as $d_t$ grows, $\tau_h$ saturates against $\varphi_{\max}$ before $\tau_l$ saturates against $\varphi_{\min}$, so the admissible band $[\tau_l,\tau_h]$ widens toward the lower side. At hard stages, advancement therefore requires near-maximal evidence whereas mild deterioration suffices to retreat. This matches the regret asymmetry argued above and yields the formal property $\mathrm{d}\tau_h/\mathrm{d}d_t \le |\mathrm{d}\tau_l/\mathrm{d}d_t|$ once $d_t$ exceeds the saturation point of $\tau_h$. All parameters are listed in Table~\ref{tab:hyperparams}.

Finally, as depicted in Eq. \eqref{dtplus1}, the difficulty adjustment policy integrates multiple performance indicators through a three-branch decision structure:
\begin{small} 
\begin{equation}
\label{dtplus1}
d_{t+1} = \begin{cases}
d_t + 1 & \text{if } \mu_w > \tau_h \land \mathcal{S}=1 \land m_t>\text{L}_{toler}, \\
d_t - 1 & \text{if } (\mu_w < \tau_l \lor \nabla^2r_t<-\text{R}_{toler}) \land m_t<-\text{L}_{toler}, \\
d_t & \text{otherwise}. 
\end{cases}
\end{equation}
\end{small}

First, the step-by-step difficulty changes ($\pm 1$) ensure gradual progression matching standard difficulty levels. Second, the strict condition of task difficulty upgrade ($\mu_w \land \mathcal{S} \land m_t$) requires concurrent high performance, stability, and positive momentum. Third, the degradation trigger ($\mu_w \lor \nabla^2r_t$) responds quickly to either direct failure or accelerating decline.
The numerical settings of $\text{L}_{toler}$ and $\text{R}_{toler}$, their physical interpretation, and the failure modes that arise when they are pushed out of range are deferred to Sec.~\ref{settings}. All FlexDiff hyperparameters are listed in Table~\ref{tab:hyperparams}.

As shown in \textbf{Algorithm} \ref{FlexDiffDP}, \textbf{FlexDiff} establishes a dynamic mapping relationship between agent performance and environmental difficulty by monitoring dual-channel metrics (i.e., \textit{Win Rate} and \textit{Episode Reward}) in real-time during MARL training, ensuring multi-dimensional learning feedback during MARL policy optimization. 

\begin{algorithm}[ht] 
\caption{The Difficulty Scheduling Process of \textbf{FlexDiff}} 
\label{FlexDiffDP}
\begin{algorithmic}[1]
\REQUIRE Observation window $\mathcal{W}_t$, current difficulty $d_t$, momentum $m_{t-1}$;
\STATE Update running statistics $\mu_w$, $\sigma_w^2$ (Eq. \ref{sigmaw2})
\STATE Compute trend slope $\beta_w$ via normal equations (Eq. \ref{betaw})
\STATE Derive reward convexity $\nabla^2 r_t$ from buffer (Eq. \ref{nabla2r})
\STATE Update momentum $m_t$ (Eq. \ref{mtupdate})
\STATE Calculate adaptive thresholds $\tau_h(d_t)$, $\tau_l(d_t)$
\STATE Difficulty Change with Eq. \ref{dtplus1}; 
\IF{$\mu_w > \tau_h$ \AND stability $\mathcal{S}=1$ \AND momentum $m_t>\text{L}_{toler}$}
\STATE Increase difficulty: $d_{t+1} \leftarrow \min(d_t + \delta_+, D_{max})$
\ELSIF{Performance collapse detected ($\mu_w < \tau_l$ \OR $\nabla^2r_t<-\text{R}_{toler}$) \AND $m_t<-\text{L}_{toler}$}
\STATE Decrease difficulty: $d_{t+1} \leftarrow \max(d_t - \delta_-, 1)$
\ENDIF
\RETURN Adjusted difficulty $d_{t+1}$;
\end{algorithmic}
\end{algorithm}

\subsubsection{Stability via Two-Timescale Separation}
\label{sec:flexdiffstability}
A natural concern is whether FlexDiff's online difficulty changes destabilise the inner CGRPA learner. They do not, because the two processes act on well-separated timescales in the sense of two-timescale stochastic approximation~\cite{borkar1997twotimescale,kondaTsitsiklis2003ActorCritic}. CGRPA updates network parameters at every gradient step (the fast timescale), whereas FlexDiff inspects $\mathcal{W}_t$ only once per evaluation cycle and, by Eq.~\eqref{dtplus1}, can change $d_t$ at most once per cycle (the slow timescale). The asymmetric thresholds $(\tau_h,\tau_l)$ together with the EMA momentum filter further enforce that, with high probability, consecutive switches are separated by at least $N$ evaluation cycles: a switch flips the sign of the directional signal, and the EMA requires $\Theta(1/(1-\gamma_m))\!\approx\!N$ cycles to cross $\pm\text{L}_{toler}$ in the opposite direction. Between switches the inner learner therefore observes a quasi-stationary MDP, so CGRPA can approach the stage-wise Bellman fixed point before the next curriculum perturbation. This is the same timescale-separation argument used to establish convergence of actor--critic schemes~\cite{kondaTsitsiklis2003ActorCritic}, and it explains why FlexDiff produces smoother learning curves than fixed-difficulty training despite operating online.

\subsection{Counterfactual Group Relative Policy Advantage (CGRPA)}
\label{sec:cgrpa}

\subsubsection{Problem and Motivation}
Consider a cooperative multi-agent system with $N$ agents modeled as a Dec-POMDP. Let $Q_{\text{tot}}(\mathbf{\tau},\mathbf{u}|s)$ denote the joint action-value function conditioned on the state $s$, where $\mathbf{\tau}=(\tau_1,...,\tau_N)$ are agents' action-observation histories and $\mathbf{u}=(u_1,...,u_N)$ the joint action. The key challenge lies in
\textit{Asymmetric Credit Assignment}, $\quad \partial Q_{\text{tot}}/\partial Q_i \geq 0$, in \textit{actor-critic} (AC)-based MARL methods (such as QMIX constraint), which could lead to sub-optimal dilemma.
Traditional value-based MARL algorithms \cite{MARL2025surveymultiagent} often struggle with policy convergence and recovery after environmental difficulty changes, especially when CL technique is applied to dynamically adjust the complexity of the task.

In light of these challenges, we introduce a novel algorithm, \textbf{CGRPA}, to enhance the agents' robustness during curriculum difficulty switching. \textbf{CGRPA} rethinks how credit is assigned to each agent’s actions by comparing their actual actions with potential counterfactual actions. This allows agents to rapidly adapt and adjust their strategies after changes in task difficulty introduced by CL.

\textit{Definition (Performance Collapse).} For the curriculum-aware learner, we declare a \textit{performance-collapse event} at evaluation step $t$ when (a) the windowed mean win rate $\mu_w$ falls below the dynamic lower threshold $\tau_l(d_t)$ \textit{or} the second-order reward difference satisfies $\nabla^2 r_t < -\text{R}_{toler}$, \textit{and} (b) the EMA momentum confirms a sustained downward trend, $m_t < -\text{L}_{toler}$. This is the same logical condition that triggers the demotion branch of Eq.~\eqref{dtplus1} and the \textsc{else if} clause of Algorithm~\ref{FlexDiffDP}; it is therefore a measurable, threshold-based event rather than a qualitative descriptor, and CGRPA's role is to keep the policy update stable across precisely these events.

\subsubsection{Action-Group Counterfactual Advantage}
\begin{figure}[ht]
\centering
\includegraphics[width=0.95\linewidth]{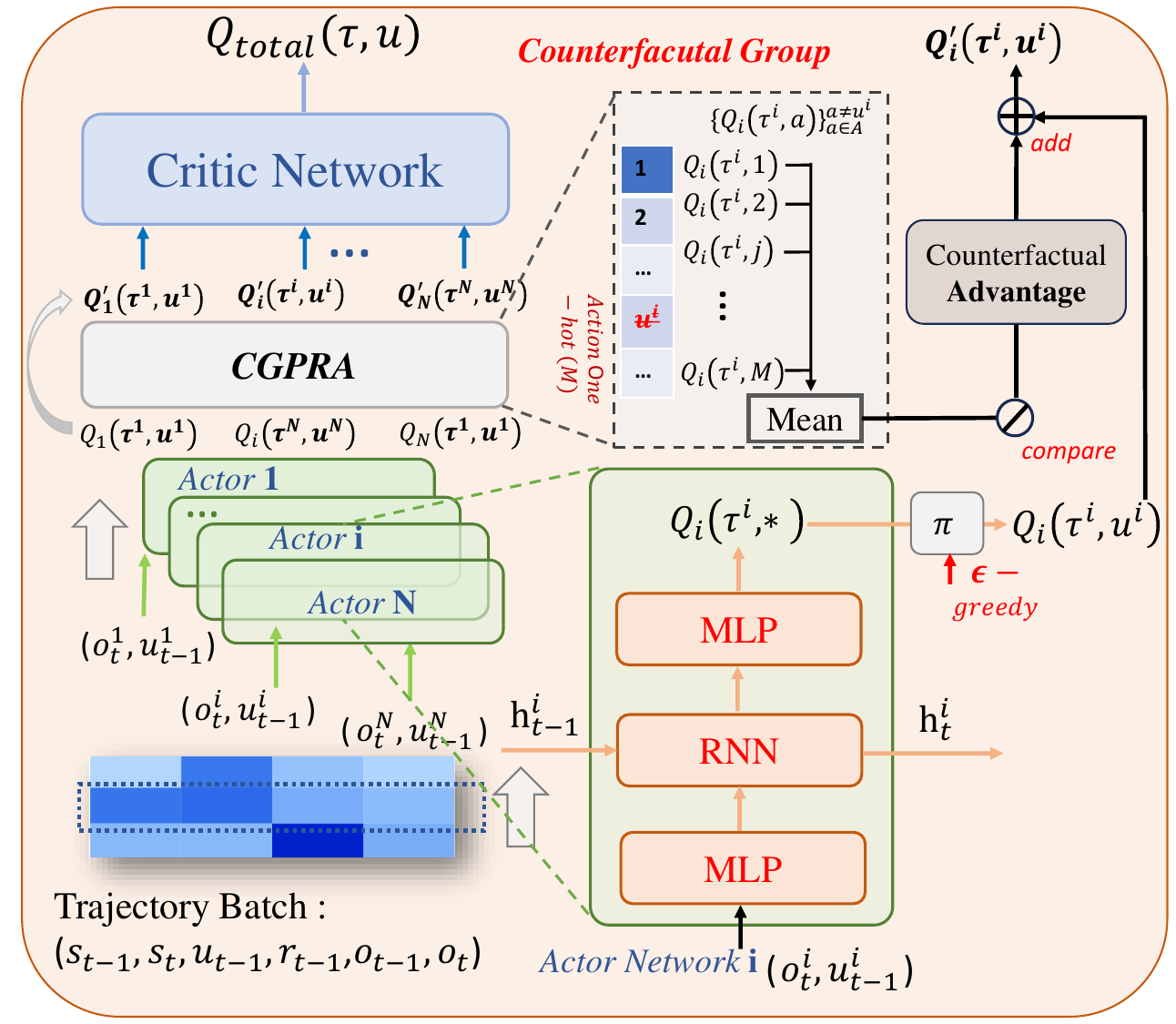} 
\caption{The detailed visualization of our CGRPA algorithm, corresponding to the Part 3 of Fig. \ref{fig_ourmodel}.} 
\label{fig_CGRPAfig}
\end{figure}

As shown in Fig. \ref{fig_CGRPAfig}, \textbf{CGRPA} reconstructs the advantage function and introduces a innovative action-group relative advantage estimation, through dual mechanisms: \textit{counterfactual reasoning} \cite{FoersterCOMA} and \textit{group relative policy optimization} \cite{shao2024deepseek} to achieve precise per-agent credit assignment.

During Q-value estimation in MARL training, the counterfactual advantage can be formalized as Eq. \eqref{eq:advantage}: 
\begin{equation*}
A^{\text{CF}}_i(s,\mathbf{u}) = \underbrace{Q_{\text{tot}}(s,\mathbf{u}) - \mathbb{E}_{\bar{u}_i\sim\pi_i}[Q_{\text{tot}}(s,(\mathbf{u}^{-i},\bar{u}_i))]}_{\text{Counterfactual Baseline}}
\end{equation*}
\begin{align}
- \alpha D_{\text{KL}}(\pi_i\|\bar{\pi}_g), \quad \quad \quad
\label{eq:advantage}
\end{align}
where $A^{\text{CF}}_i(s,\mathbf{u})$ denotes the counterfactual advantage of agent $i$ at state $s$ under joint action $\mathbf{u}=(u_1,\dots,u_N)$, $Q_{\text{tot}}(s,\mathbf{u})$ is the joint action-value function produced by the centralized mixing network, $\mathbf{u}^{-i}$ represents the joint actions excluding the $i$-th agent's action, $\bar{u}_i\sim\pi_i$ is a counterfactual action sampled from agent $i$'s own policy $\pi_i$, $\bar{\pi}_g(u|s) = \frac{1}{N}\sum_{j=1}^N\pi_j(u|\tau_j)$ represents the group-averaged policy, $D_{\text{KL}}(\pi_i\|\bar{\pi}_g)$ measures the Kullback--Leibler divergence between agent $i$'s policy and the group-averaged policy, and $\alpha>0$ is a temperature coefficient controlling the coordination strength\footnote{The KL divergence $D_{\text{KL}}$ in Eq. \eqref{eq:advantage} creates a potential game where agents are rewarded for aligning with $\bar{\pi}_g$.}.
The first term quantifies individual contribution by comparing the actual action $u_i$ with its policy's averaged expectation value, while the \textit{KL divergence} term, $D_{\text{KL}}$, constrains policy deviation from the group average. The temperature parameter $\alpha$ balances these components. This design breaks traditional valuable-based MARL methods' (such as VDN, QMIX) monotonicity constraint, enabling asymmetric and effective credit assignment.

\subsubsection{Group-Relative Policy Optimization}
The policy update rule combines counterfactual credit assignment with group relative policy:
\begin{align}
\nabla_\theta J(\theta) &= \mathbb{E}_{\mathcal{D}}\left[\sum_{i=1}^N A^{\text{CF}}_i(s,\mathbf{u}) \nabla_\theta\log\pi_i(u_i|\tau_i)\right] \\
\text{s.t.}\quad &Q_{\text{tot}} = \text{Critic}(Q_1,...,Q_N;\psi),\ \psi\in\Psi_{\text{mono}}.
\label{eq:policy_update}
\end{align}
where $\nabla_\theta J(\theta)$ denotes the gradient of the objective with respect to the joint policy parameters $\theta$, $\mathbb{E}_{\mathcal{D}}[\cdot]$ represents the expectation over trajectories sampled from the replay buffer $\mathcal{D}$, $\pi_i(u_i|\tau_i)$ is the policy of the $i$-th agent conditioned on its local action-observation history $\tau_i$, $\bar{\pi}_g(u|s) = \frac{1}{N}\sum_{j=1}^N\pi_j(u|\tau_j)$ represents the group-averaged policy, $\text{Critic}(Q_1,\dots,Q_N;\psi)$ is the centralized mixing network parameterized by $\psi$, and $\Psi_{\text{mono}}$ denotes the constraint set enforcing monotonicity (i.e., $\partial Q_{\text{tot}}/\partial Q_i \geq 0$ for all $i$).

Typically, there are two primary distinctions from conventional AC-based MARL frameworks. First, during execution, each $\pi_i$ relies solely on local history $\tau_i$, thereby preserving decentralization. 
Second, the training process utilizes a critic network (e.g., the mixing network in QMIX \cite{QMIX}), denoted as $\text{Critic}(\cdot;\psi)$, which theoretically preserves monotonicity while actively reducing projection bias through counterfactual advantage-weighted learning.

Specifically, the $A^{\text{CF}}$ term in Eq. \eqref{eq:advantage} guides gradient updates through two mechanisms. The counterfactual component enables more effective individual contribution assessment, while the KL divergence maintaining policy consistency. 

\begin{algorithm}[ht]
\caption{Implementation Details of CGRPA}
\label{alg:cgrpa}
 \small
 \begin{algorithmic}[1]
     \REQUIRE Learning rate $\eta_{\text{LR}}$, discount factor $\gamma$, critic network $f_\psi$
 \STATE Initialize: Agent networks $\{Q_i, \pi_i\}_{i=1}^N$ with parameters $\theta$, Target networks $\{Q_i^-, \pi_i^-\}_{i=1}^N \leftarrow \{Q_i, \pi_i\}$, Critic networks $f_\psi$ and target $f_{\psi^-} \leftarrow f_\psi$, Replay buffer $\mathcal{D} \leftarrow \emptyset$.

 \FOR{episode $=1$ \TO $M$}
     \STATE Collect trajectory $\{(s_t,\bm{\tau}_t,\bm{u}_t,r_t)\}$ using $\epsilon$-greedy;
     \FOR{gradient step $=1$ \TO $K$}
         \STATE Sample mini-batch $B \sim \mathcal{D}$;
         \STATE Compute counterfactual advantage $A^{\text{CF}}_i$; (Eq. \ref{eq:advantage})
         \STATE Compute new Q-values for each $i-th$ agent, where $i\in \{1, \dots, N\}$; (Eq. \ref{Qi}): 
         \begin{equation*}
             \tilde{Q}_i = Q_i(\tau_i,u_i) + \lambda A^{\text{CF}}_i
         \end{equation*}
        
         \STATE Calculate current $Q_{tot}$: $Q_{\text{tot}} = f_\psi\left(\{\tilde{Q}_i\}, s\right)$;
        
         \STATE Compute target actions $\bm{u}'^*$:
         \begin{equation*}
         \bm{u}'^* = argmax_{\bm{u}'}f_{\psi^-}(\{Q_i^-(\tau_i',u_i')\}, s');
         \end{equation*}
        
         \STATE Calculate target value:
         \begin{equation*}
             y = r + \gamma \cdot f_{\psi^-}\left(\{Q_i^-(\tau_i',u_i'^*) + \lambda A^{\text{CF}}_i\}, s'\right);
         \end{equation*}
        
         \STATE Update mixing network via:
         \begin{equation*}
             \psi \leftarrow \psi - \eta_{\text{LR}}\nabla_\psi \frac{1}{|B|}\sum_{k=1}^{|B|} (Q^{k}_{\text{tot}} - y^{k})^2;
         \end{equation*}

         \STATE Update agent policies with coordination regularization:
         \( \theta_i \leftarrow \theta_i - \eta_{\text{LR}}\nabla_{\theta_i} \Big[ A^{\text{CF}}_i \log\pi_i(u_i|\tau_i) + \alpha D_{\text{KL}}  \) \(\left(\pi_i\|\bar{\pi}_g\right) \Big] \quad \forall \text{ agent } i.\)
     \ENDFOR
 \ENDFOR
 \end{algorithmic}
\end{algorithm}

\subsubsection{Q-Function Update with Counterfactual Infusion}
We incorporate the counterfactual advantages $A^{\text{CF}}_i(s,\bm{u})$ into individual Q-value estimation: 
\begin{align}
\label{Qi}
\tilde{Q}_i(\tau_i,u_i) &= Q_i(\tau_i,u_i) + \lambda A^{\text{CF}}_i(s,\bm{u}),
\end{align}
\begin{align}
\label{Qtot}
Q_{\text{tot}} &= f_{\psi}\left(\{\tilde{Q}_i\}_{i=1}^N, s\right),
\end{align}
where $\tilde{Q}_i(\tau_i,u_i)$ is the counterfactual-augmented individual Q-value of agent $i$, $Q_i(\tau_i,u_i)$ denotes its original local Q-value evaluated on history $\tau_i$ and action $u_i$, $\lambda \in (0,1]$ controls the counterfactual-integration strength, $A^{\text{CF}}_i(s,\bm{u})$ is the counterfactual advantage from Eq. \eqref{eq:advantage}, $f_{\psi}(\cdot,s)$ is the centralized mixing network that aggregates $\{\tilde{Q}_i\}_{i=1}^N$ into the joint value $Q_{\text{tot}}$ at state $s$ under parameters $\psi$, and the monotonicity property $\partial f_\psi / \partial \tilde{Q}_i \geq 0$ is preserved by construction.

The modified TD-target computation becomes:
\begin{equation}
y = r + \gamma f_{\psi^-}\left(\{Q_i^-(\tau_i',u_i'^*) + \lambda A^{\text{CF}}_i(s',\mathbf{u}'^*)\}, s'\right),
\end{equation}
where $y$ is the temporal-difference target, $r$ denotes the immediate global reward, $\gamma \in [0,1)$ is the discount factor, $s'$ and $\tau_i'$ represent the next state and the next action-observation history of agent $i$, $\mathbf{u}'^* = (u_1'^*,\dots,u_N'^*)$ is the joint greedy action selected by the target networks, $\psi^-$ and $Q_i^-$ are the parameters of the target mixing network $f_{\psi^-}$ and target individual Q-networks (slowly tracking $\psi$ and $Q_i$), and the bracket term $\{\cdot\}$ collects each agent's counterfactual-augmented next-step Q-value. The complete loss function is:
\begin{equation}
\mathcal{L}(\theta) = \mathbb{E}_{\mathcal{D}}\left[\left(f_\psi(\{\tilde{Q}_i\},s) - y\right)^2\right] + \beta \sum_{i=1}^N D_{\text{KL}}(\pi_i\|\bar{\pi}_g),
\end{equation}
where $\mathcal{L}(\theta)$ denotes the joint training objective over the policy and critic parameters $\theta$, $\mathbb{E}_{\mathcal{D}}[\cdot]$ is the expectation taken over mini-batches drawn from the replay buffer $\mathcal{D}$, the first term is the standard TD squared-error between the predicted $Q_{\text{tot}}$ and the bootstrap target $y$, $\beta>0$ is the weight of the policy-coordination regularizer, and $D_{\text{KL}}(\pi_i\|\bar{\pi}_g)$ measures the divergence between agent $i$'s policy and the group-averaged policy $\bar{\pi}_g$ defined in Eq.~\eqref{eq:policy_update}.


For simplicity, Algorithm~\ref{alg:cgrpa} outlines the key steps of our \textbf{CGRPA} method. Overall, \textbf{CGRPA} improves credit assignment by comparing each agent's actual actions with potential counterfactual actions, allowing agents to quickly adapt their strategies to the task complexity variations introduced by CL.


\begin{table*}[ht]
\renewcommand{\arraystretch}{1.05}
\caption{Details of the adopted SMAC experimental scenarios and maps, covering Easy, Hard, and Super-Hard levels.}
\label{table_smacmaps}
    \footnotesize 
\centering
\scalebox{0.825}{
\begin{tabular}{ccccc@{\hspace{2em}}ccccc}
\cmidrule(lr){1-5} \cmidrule(lr){6-10}
\textbf{Map Name} & \textbf{Difficulty} & \textbf{State Dim} & \textbf{Action Dim} & \textbf{Ep. Len} &
\textbf{Map Name} & \textbf{Difficulty} & \textbf{State Dim} & \textbf{Action Dim} & \textbf{Ep. Len} \\
\cmidrule(lr){1-5} \cmidrule(lr){6-10}
3m & Easy & 90 & 13 & 60 &
5m & Easy & 135 & 15 & 120 \\
8m & Easy & 192 & 18 & 120 &
2s\_3z & Easy & 130 & 12 & 120 \\
1c\_3s\_5z & Easy & 270 & 15 & 180 &
3s\_5z & Easy & 216 & 14 & 150 \\
\cmidrule(lr){1-5} \cmidrule(lr){6-10}
5s10z & Hard & 380 & 25 & 200 &
7s7z & Hard & 350 & 24 & 200 \\
5m\_vs\_6m & Hard & 98 & 12 & 70 &
8m\_vs\_9m & Hard & 154 & 14 & 120 \\
3s\_vs\_5z & Hard & 68 & 11 & 250 &
2c\_vs\_64zg & Hard & 396 & 65 & 600 \\
bane\_vs\_bane & Hard & 288 & 25 & 200 &
1c3s8z\_vs\_1c3s9z & Hard & 425 & 33 & 180 \\
\cmidrule(lr){1-5} \cmidrule(lr){6-10}
MMM2 & Super Hard & 322 & 18 & 180 &
3s5z\_vs\_3s6z & Super Hard & 230 & 15 & 170 \\
27m\_vs\_30m & Super Hard & 342 & 31 & 180 & & & & & \\
\cmidrule(lr){1-5} \cmidrule(lr){6-10}
\end{tabular}
}
\end{table*}

\section{Experiments}
This section validates CL-MARL through systematic experiments. Section \ref{expersetup} details the setup, including the SMAC benchmark, hyperparameters, and hardware. Section \ref{mainres} compares CL-MARL with baselines across SMAC maps of varying difficulty. Section \ref{ablares} presents ablation studies on component importance, performance gains, and theoretical support.

\begin{figure}[ht]
\centering
\includegraphics[width=0.95\linewidth]{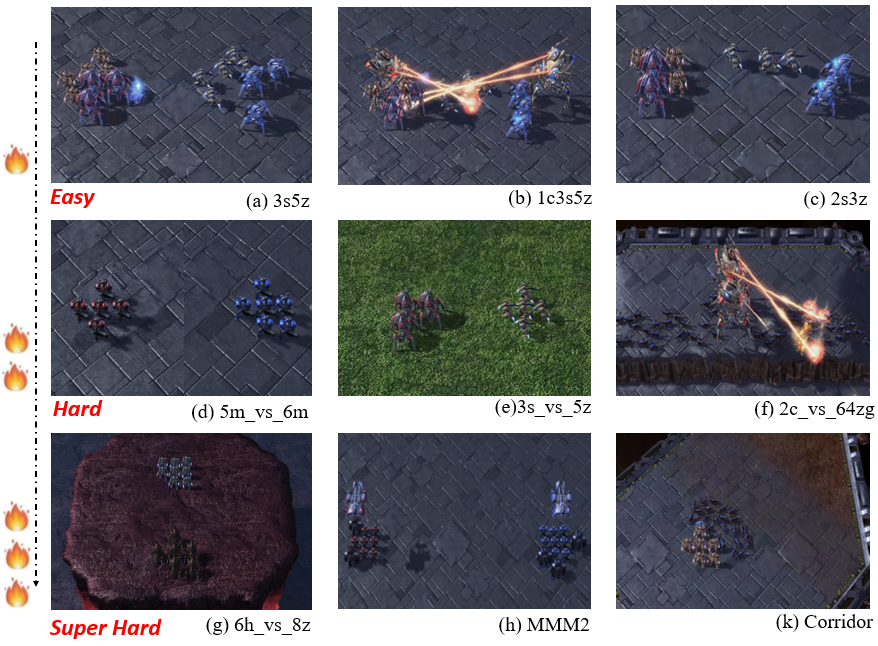}
\caption{The typical SMAC benchmark scenarios and maps with progressive difficulty levels, ranging from \textit{Easy}, \textit{Hard} to \textit{Super-Hard}, for MARL algorithm testing.}
\label{fig_SMACMAPS}
\end{figure}

\subsection{Experimental Setup}
\label{expersetup}
This section details the experimental setup, including: the adopted SMAC environment, model hyperparameter configurations, experimental procedure designing, and hardware implementation specifications.

\subsubsection{Decentralised StarCraft II Micromanagement}
\label{smacenviron}
The SMAC environment is a decentralized MARL benchmark based on \textit{StarCraft II} micromanagement \cite{ellis2023smacv2}. In SMAC, each RL agent controls an individual unit (Marines, Stalkers, etc.) with partial observability [\texttt{local unit positions}, \texttt{health}] and discrete actions [\texttt{move}, \texttt{attack}, \texttt{stop}]. 
On the allied side, each unit is independently controlled by decentralized learning agents, contrasting with enemy units that are managed by StarCraft II's native AI utilizing predefined heuristic rules.
The action space of SMAC agents is discrete, comprising \textit{movement} [4 directions], \textit{attack} [if target is in range], \textit{stop}, and \textit{no-op} commands. Each agent receives partial observations including relative positions, distances, types, and status (health, shields) of units within its sight range. During training, a global state (hidden during testing) provides full information about all units' coordinates, health, cooldowns, and allies' previous actions.
The detailed configurations of typical SMAC cooperative scenarios across different difficulty levels are provided in Table \ref{table_smacmaps}. 
In this table, \textit{State Dim} representing the state space dimension, \textit{Action Dim} indicating the action space dimension, and \textit{Ep. Len} denotes the episode length measured in time steps. 
Notably, Super Hard scenarios like \texttt{3s5z\_vs\_3s6z} and \texttt{corridor} remain unsolved by SOTA methods, such as QMIX \cite{QMIX}, MARR \cite{MARR}, DER \cite{DERhu2023discriminative}, EMC \cite{EMCzheng2021episodic} and EMU \cite{emu2024}, establishing SMAC as a key benchmark for cooperative MARL research.

\begin{table*}[ht]
\renewcommand{\arraystretch}{1.05}
\caption{Experimental Hyperparameter Settings for our CL-MARL Framework within QMIXBackbone}
    \label{tab:hyperparams}
    \centering
                \scalebox{0.8}{
\begin{tabular}{lclclclc}
\hline
\makecell{\textbf{Parameters} \\ \textbf{(ori-QMIX)}} & \textbf{Value} & \makecell{\textbf{Parameters} \\ \textbf{(CL-MARL)}} & \textbf{Value} & \makecell{\textbf{Parameters} \\ \textbf{(ori-QMIX)}} & \textbf{Value} & \makecell{\textbf{Parameters} \\ \textbf{(CL-MARL)}} & \textbf{Value} \\
\cmidrule(lr){1-4} \cmidrule(lr){5-8}
Learning rate ($\eta_{\text{LR}}$) & 5e-4 & Maximum Difficulty & 10 & RNN hidden dim & 64 & Start Difficulty & 5 \\
Gradient norm clip & 10 & Minimum Difficulty & 1 & QMIX hidden dim & 32 & sliding window length & 20 \\
Discount factor ($\gamma$) & 0.99 & Default Difficulty & 7 & Training Batch size & 32 & win-rate increase limit & 0.08 \\
\makecell{Anchored Difficulty \\ Level (Diff$_{anch}$)} & 5 & \makecell{Scale \\ coefficient ($\text{Scal}_{coef}$)} & 0.02 & Replay buffer capacity & 5,000 & reward increase limit & 0.1 \\
\makecell{$\text{L}_{toler}$, Tolerance Level \\ of momentum, $m_t$} & 0.2 & \makecell{Tolerance Reward \\ ($\text{R}_{toler}$)} & 0.05 & Target update interval & 200 & momentum threshold $\zeta$ & 0.1 \\
initial $\epsilon$ & 1.0 & \makecell{max win-rate $\varphi_{\max}$ \\ (FlexDiff)} & 0.75 & minimized $\epsilon$ & 0.05 & \makecell{min win-rate $\varphi_{\min}$ \\ (FlexDiff)} & 0.25 \\
$\epsilon$-anneal steps & 50,000 & CGRPA Loss Weight $\lambda$ & 0.5 & & & & \\ 
\hline
\end{tabular}
    }
\end{table*}

SMACv2\footnote{SMACv2 Code: \url{https://github.com/oxwhirl/smacv2}.} \cite{ellis2023smacv2} enhances the original SMAC benchmark through three initialization-phase modifications while preserving the fundamental map configurations.
Across all scenarios, it introduces randomized unit starting positions (either in \textit{surrounded} or \textit{reflected} configurations) and adjusts sight and attack ranges to align with official StarCraft II values. Additionally, for several specific and more challenging scenarios, SMACv2 introduces randomized army compositions, where unit types and quantities are sampled from a controlled pool for each race (\textit{Terran, Protoss, and Zerg}), preventing agents from overfitting to fixed strategies.
These modifications enhance scenario diversity and prevent algorithmic overfitting. Notably, the built-in opponent difficulty API in SMACv2 serves as an effective difficulty measurer in our CL-MARL, making SMACv2 particularly suitable for progressive training. 
Thus, the experiments are entirely based on the enhanced \textit{StarCraft-II} benchmark, SMACv2. 

To comprehensively validate the effectiveness of our propoed method, we conduct extensive experiments across most SMAC scenarios, spanning three difficulty levels: \textit{Easy}, \textit{Hard}, and \textit{Super Hard}. Through these thorough evaluations, we demonstrate the universal superiority of our CL-MARL framework. Meanwhile, we list several representative scenarios from each difficulty level in Fig. \ref{fig_SMACMAPS}.

\begin{figure*}[ht]
\centering
\includegraphics[width=0.95\linewidth]{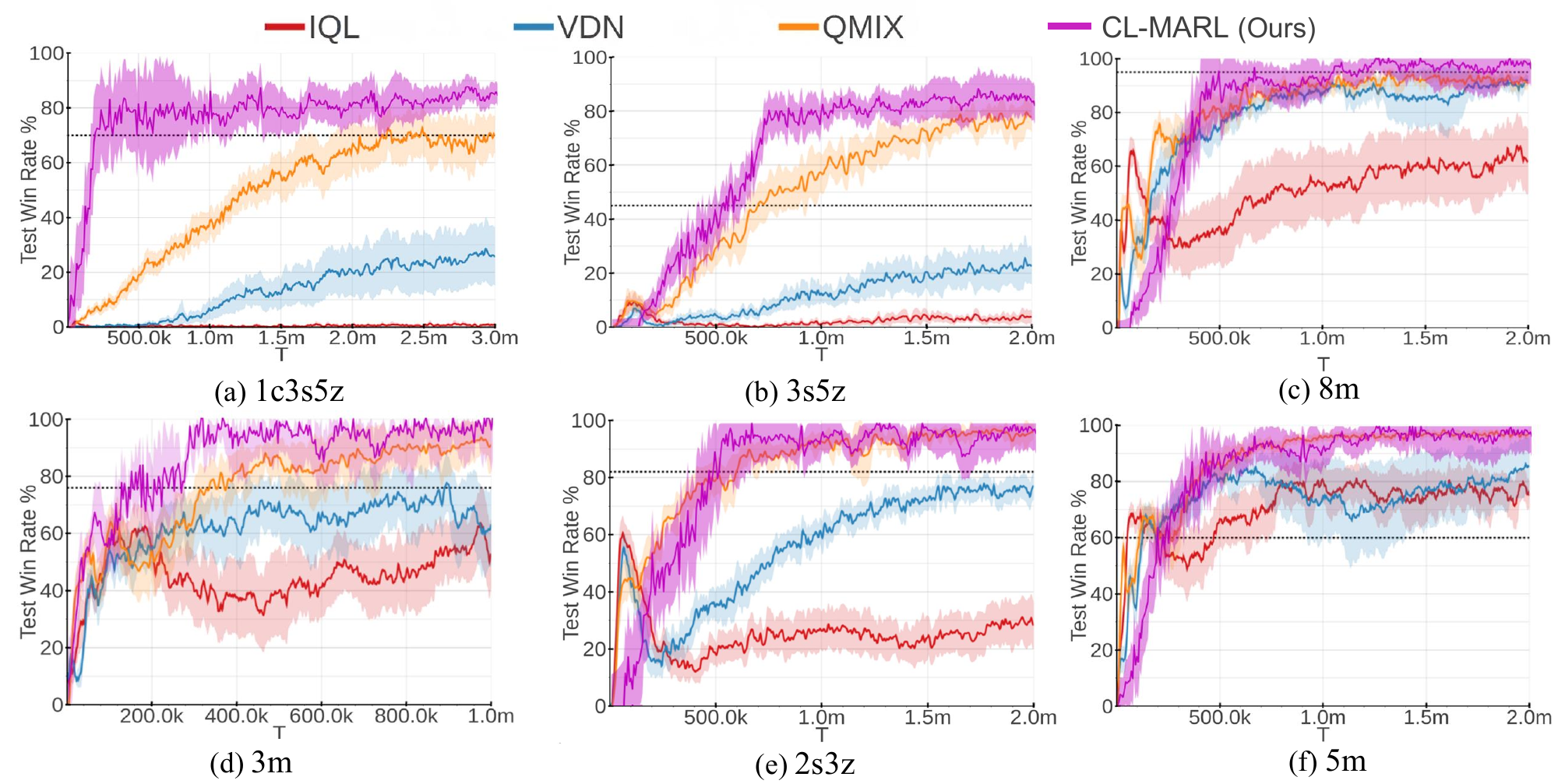}
\caption{Baseline Comparisons of the six easy maps in the SMAC benchmark. The x-axis represents the global training steps, while the y-axis records the average win rate progression of the RL algorithm.}
\label{fig_easyresults}
\end{figure*}

\subsubsection{Architecture and Training Details}
\label{settings}
Our proposed CL-MARL framework modifies the original QMIX \cite{QMIX} architecture while maintaining identical hyperparameter configurations to ensure the overall experimental rigor and comparability
\footnote{The source codes of QMIX and other MARL algorithms are available at \url{https://github.com/starry-sky6688/MARL-Algorithms}}.
The hyperparameter configurations for the original QMIX backbone network and the newly introduced CL-MARL modules, are detailed in Table \ref{tab:hyperparams}. 
The full implementation details, including environment configuration, network architecture, and training procedures, can be found in our released project\footnote{Our code of CL-MARL: \url{https://github.com/NICE-HKU/CL2MARL-SMAC}.}.

The evaluation metric in SMAC is the \textit{test-win-rate}, which is measured at different training stages\footnote{The win-rate is calculated by the fraction of test episodes in which the learning agents successfully destroy all enemy units within the time limit.}. 
All experimental results, i.e., mean test-win-rates, reported in subsequent sections are computed over 5 independent runs for each method across selected maps, accompanied by 85\% confidence intervals. 
The implementation of CL-MARL is based on PyTorch 1.12.1
and Python 3.8.5,
running on Ubuntu 20.04.1. The hardware platform consists of four Intel Xeon Platinum 8488C CPUs (48 cores each) and four NVIDIA GeForce RTX 4090 GPUs.

\paragraph*{FlexDiff hyperparameter selection}
Beyond the QMIX backbone, FlexDiff introduces several rule-design hyperparameters whose values are listed in Table~\ref{tab:hyperparams}. We summarise their selection rationale below; the rule-level ablations in Sec.~\ref{sec:abla_rules} provide the empirical evidence supporting these choices.

\textit{Momentum threshold $\zeta$ (Eq.~\eqref{betaw}).} $\zeta$ should be the smallest slope statistically distinguishable from sampling noise. For a window of length $N$ with per-evaluation win-rate standard deviation $\sigma_w$, the standard error of an OLS slope estimate scales as $\sigma_w\sqrt{12/(N(N^2-1))}$; with $N{=}20$ and the empirical $\sigma_w\!\approx\!0.07$ observed during mid-curriculum training, a one-sigma detectable slope is $\approx0.05$ per evaluation cycle. We therefore use $\zeta{=}0.1$, roughly two standard errors above noise, and recommend $\zeta\!\in\![0.05,0.15]$ for $N\!\in\![10,30]$. Setting $\zeta$ too small yields false-positive promotions on noisy upticks; too large suppresses genuine learning signals and slows curriculum progression.

\textit{EMA decay $\gamma_m$ (Eq.~\eqref{mtupdate}).} The effective memory horizon is $\approx 1/(1-\gamma_m)$ evaluation cycles. We use $\gamma_m{=}0.9$, giving a horizon of about ten cycles. Values below $0.7$ make $m_t$ track noise, while values above $0.95$ delay reaction to genuine collapses.

\textit{Threshold magnitudes $\text{L}_{toler}$, $\text{R}_{toler}$ (Eq.~\eqref{dtplus1} and Algorithm~\ref{FlexDiffDP}).} The momentum gate $|m_t|>\text{L}_{toler}{=}0.2$ requires the EMA-filtered trend signal to exceed one fifth of its $[-1,1]$ dynamic range, which after $\tanh$ saturation corresponds to a sustained directional move rather than a single-cycle fluctuation. Setting $\text{L}_{toler}<0.1$ re-introduces chattering, whereas $\text{L}_{toler}>0.3$ effectively freezes the curriculum. The reward-convexity floor $\nabla^2 r_t<-\text{R}_{toler}$ with $\text{R}_{toler}{=}0.05$ in normalised reward units targets roughly a $5\%$ per-cycle acceleration of decline, which we found to be the smallest second-order signal that consistently precedes win-rate collapse on SMAC. Raising $\text{R}_{toler}$ delays demotion until damage has accumulated, while lowering it triggers spurious retreats. The negative-momentum confirmation $m_t<-\text{L}_{toler}$ prevents demotion from being triggered by an isolated $\nabla^2 r_t$ outlier.

\paragraph*{On the size of the hyperparameter set}
We acknowledge that FlexDiff carries a relatively large set of hyperparameters compared with the vanilla QMIX backbone. The values reported in Table~\ref{tab:hyperparams} were obtained through systematic search on a held-out subset of SMACv2 maps and represent the best configurations we found; the rule-level ablations in Sec.~\ref{sec:abla_rules} further confirm that performance is robust within the recommended ranges. Looking ahead, a more automated solution is desirable. In future work we plan to introduce an adaptive tuning scheme that lets the model adjust the curriculum-learning hyperparameters online during training -- for instance, by formulating $\zeta$, $\gamma_m$, and $(\text{L}_{toler}, \text{R}_{toler})$ as a small meta-learning problem optimised against the same dual-metric signals used by FlexDiff -- so as to reduce the need for manual calibration and improve cross-domain generality.

\subsection{Main Comparison Results}
\label{mainres}
All experiments reported below are run on the \textit{SMACv2} benchmark; map names follow the SMAC convention by historical practice, but unit start positions and sight ranges are randomised per the SMACv2 defaults, and randomised army compositions are enabled on the scenarios that support them. Whenever we refer to an environment by a SMAC-style label (e.g., \texttt{8m}, \texttt{MMM2}), the underlying configuration is the SMACv2 variant of that scenario.
In this section, we present extensive experimental results on the SMAC benchmark to validate the effectiveness and superiority of our proposed CL-MARL framework in the multi-agent cooperative-adversarial scenarios.

We systematically compare our method against SOTA MARL algorithms, including VDN \cite{VDN}, QMIX \cite{QMIX}, OW-QMIX \cite{WeightedQMIX}, QTRAN \cite{QTRAN}, QPLEX \cite{wang2021qplex}, MARR \cite{MARR}, DER \cite{DERhu2023discriminative}, EMC \cite{EMCzheng2021episodic}, and EMU \cite{emu2024}, across a diverse set of SMAC maps spanning three difficulty levels: \textit{Easy, Hard}, and \textit{Super Hard}.
In this study, we evaluated our method on nearly 20 different SMAC benchmark maps, significantly more than in previous works, such as QMIX and OW-QMIX \cite{WeightedQMIX,QMIX} (6 maps each), DER \cite{DERhu2023discriminative} (4 maps), and MARR \cite{MARR} (7 maps), EMC \cite{EMCzheng2021episodic} (11 maps), EMU \cite{emu2024} (10 maps).
To the best of our knowledge, our experiment represents one of the most comprehensive evaluations to date on SMAC benchmark, encompassing a wide variety of situations, from simple skirmishes to highly complex large-scale battles. 

Based on the average task difficulty of maps tested, we organize our experimental comparisons into three parts: Sections \ref{easysubmain}, \ref{hardersubmain}, and \ref{hardestsubmain}. Since our CL-MARL is built based on the original QMIX \cite{QMIX}, the subsequent experiments will focus more on comparisons between our model and QMIX as well as its related variants \cite{WeightedQMIX,QTRAN,wang2021qplex}. 

\begin{figure*}[htbp]
\centering
\includegraphics[width=0.985\linewidth]{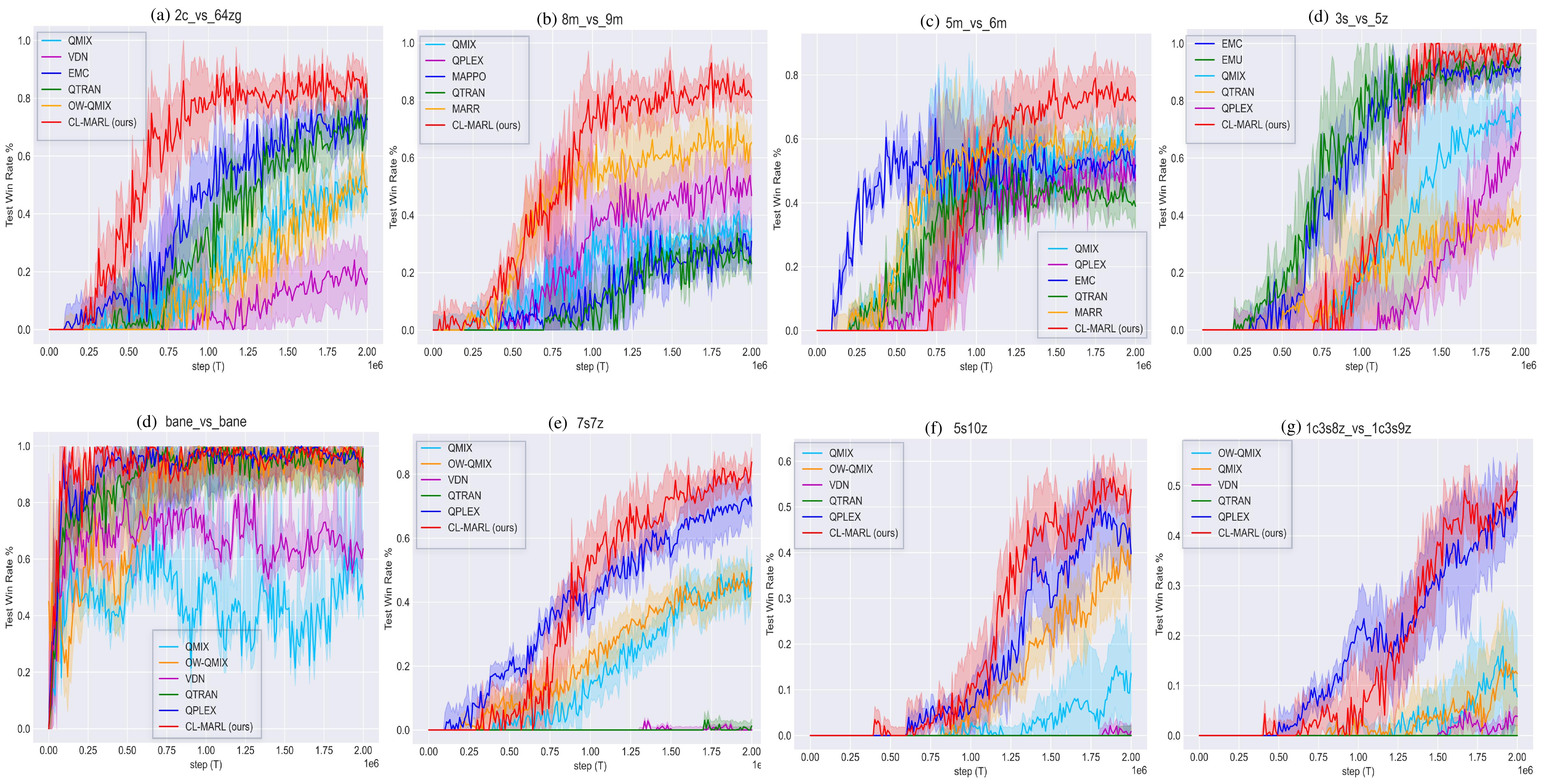}
\caption{Baseline Comparisons of the eight maps of \textit{Hard Level} in the SMAC benchmark [\textit{x-axis}: global training steps, \textit{y-axis}: average win rate].}  %
\label{fig_hardresults}
\end{figure*}

\subsubsection{Performance on Easy Maps}
\label{easysubmain}
First, we evaluate the learning capability of our proposed CL-MARL on six easy maps. As shown in Fig. \ref{fig_easyresults}, the test maps are: (a) \texttt{1c3s5z}, (b) \texttt{3s5z}, (c) \texttt{8m}, (d) \texttt{3m}, (e) \texttt{2s3z}, and (f) \texttt{5m}. The map names indicate the number and types of units for both the learning agents and their opponents. For instance, \texttt{2s\_vs\_1sc} denotes 2 \textit{Stalkers} (controlled by RL agents) versus 1 \textit{Spine Crawler} (controlled by the SC2 script), and \texttt{1c3s5z} represents both adversarial sides having the unit composition, i.e., the agents of \{1 \textit{Colossus}, 3 \textit{Stalkers}, and 5 \textit{Zealots}\}.

Across these maps, CL-MARL consistently demonstrates strong performance in learning optimal joint policies.
Specifically, CL-MARL achieves a 100\% win rate on four maps and demonstrates significantly faster policy exploration speed on the \texttt{3m}, \texttt{3s5z}, and \texttt{1c3s5z} maps.
Notably, in more complex maps like \texttt{1c3s5z} and \texttt{3s5z}, conventional algorithms such as IQL, VDN, and QMIX often stagnate in local optima, likely due to the static difficulty setting.
In contrast, CL-MARL consistently avoids these suboptimal solutions by leveraging progressive difficulty adaptation. Furthermore, on maps like \texttt{2s3z} and \texttt{8m}, CL-MARL converges to the optimal policy notably earlier than other approaches.
This improvement is attributed to the CL-based framework, which enables agents to acquire strategies in a structured, easy-to-hard progression and generalize more effectively across tasks.

\begin{figure*}[ht]
\centering
\includegraphics[width=0.85\linewidth]{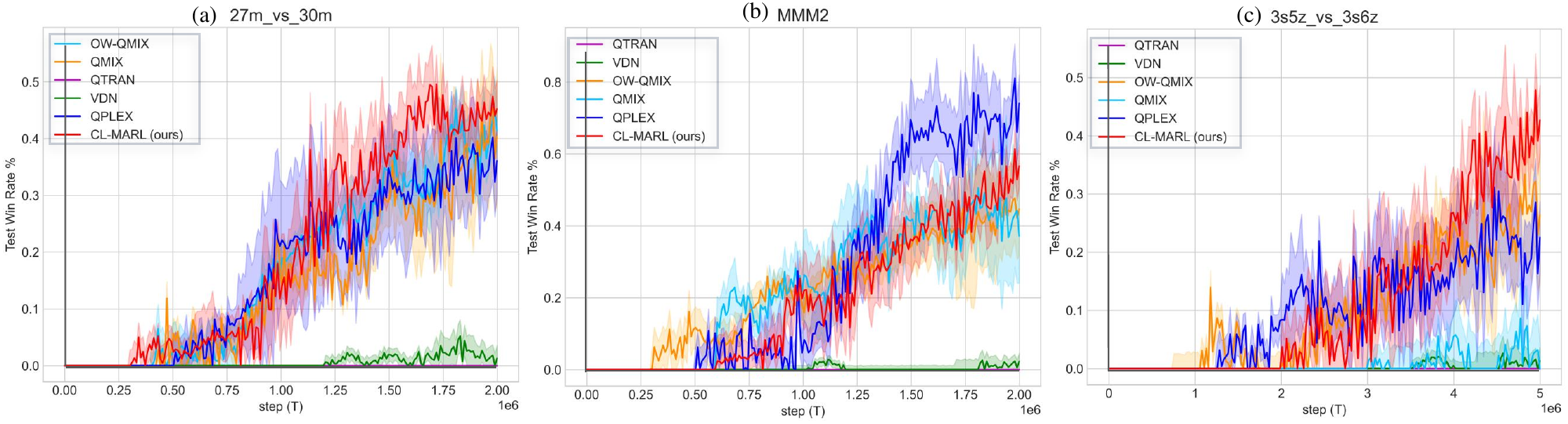}
\caption{Baseline Comparisons of the eight maps of \textit{Super Hard Level} in the SMAC benchmark [\textit{x-axis}: global training steps, \textit{y-axis}: win rate].} 
\label{fig_hardestresults}
\end{figure*}

While Fig. \ref{fig_easyresults} highlights improvements in both win rate and convergence speed, we observe that CL-MARL occasionally exhibits less training stability than baseline algorithms (QMIX and VDN) during phases of rapid performance gain. This transient instability arises from the \textbf{FlexDiff} module's dynamic adjustment of task difficulty, which may temporarily perturb performance consistency.
However, this effect is limited and does not compromise final performance.
More importantly, these observations validate that our proposed \textbf{CGRPA} module ensures robust policy adaptation during difficulty transitions, effectively mitigating instability and preserving long-term learning efficiency.

\subsubsection{Performance on Hard Maps}
\label{hardersubmain}
Fig. \ref{fig_hardresults} presents the experimental results of our CL-MARL on eight challenging maps. The results clearly demonstrate that our method achieves comparable performance with SOTA baseline algorithms in most scenarios, including \texttt{2c\_vs\_64zg}, \texttt{8m\_vs\_9m}, and \texttt{5m\_vs\_6m}.
These findings further validate the effectiveness of our approach in high-difficulty environments. Notably, CL-MARL outperforms the current SOTA algorithm EMC (a curiosity-driven episodic memory variant of QPLEX) by over 8\%-14\% in win rate on the \texttt{2c\_vs\_64zg} map. Similarly, it surpasses another SOTA algorithm MARR by 10\%-13\% on both \texttt{8m\_vs\_9m} and \texttt{5m\_vs\_6m} maps.

As a enhanced variant of QMIX, CL-MARL significantly outperforms the original QMIX across all challenging maps while maintaining identical hyperparameters. For instance, on the \texttt{7s7z}, \texttt{5s10z}, and \texttt{1c3s8z\_vs\_1c3s9z} maps, the original QMIX achieves only approximately 45\%, 10\%, and 15\% win rates after total 2 million training steps, respectively. In contrast, our CL-MARL demonstrates significant improvements, boosting the final win rates by approximately 20\%, 40\%, and 35\% on these maps respectively.
Moreover, CL-MARL maintains competitive performance compared to other SOTA algorithms including QTRAN, OW-QMIX, and QPLEX. 
These substantial gains underscore the effectiveness of our CL framework in enabling robust policy learning under difficult conditions, without requiring any parameter tuning. 

Moreover, the results shown in Fig. \ref{fig_hardresults} reveal that CL-MARL exhibits slower initial convergence speed compared to baselines on certain maps, including \texttt{7s7z}, \texttt{5m\_vs\_6m}, and \texttt{3s\_vs\_5z}. While baseline algorithms typically begin showing win rate improvements around 200,000 steps, CL-MARL may require additional training steps over 500,000 steps during the early phase to adapt to varying environmental difficulties. 
This phenomenon primarily arises from the discrepancy between the initial training difficulty levels and the default testing difficulty settings.
However, this temporary convergence delay does not compromise the final performance of our CL-MARL. The CL style in CL-MARL follows a progressive reinforcement process from simple to complex tasks. This design ensures robust learning in early stages, which ultimately contributes to superior capability in discovering optimal joint strategies. Overall, the initial learning stability trade-off is effectively compensated by the algorithm's enhanced final performance and better generalization ability.

\begin{figure*}[ht]
\centering
\includegraphics[width=0.85\linewidth]{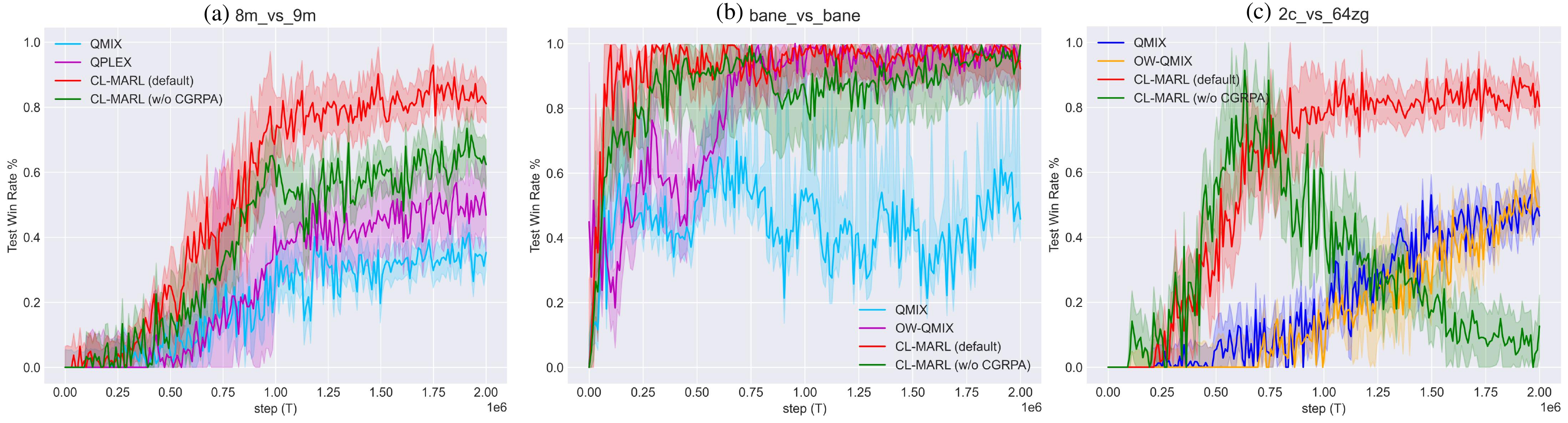} 
\caption{Performance Comparison of Convergence Risks Induced by \textbf{FlexDiff} and Stabilization Effects of \textbf{CGRPA} on CL-MARL Learning.}
\label{fig_abla1fig}
\end{figure*}

\subsubsection{Performance on Super Hard Maps}
\label{hardestsubmain}
The experiments on three super hard SMAC maps (\texttt{27m\_vs\_30m}, \texttt{MMM2}, and \texttt{3s5z\_vs}) \_3s6z reveals distinct performance characteristics of CL-MARL under extreme challenges, as shown in Fig. \ref{fig_hardestresults}.

In the \texttt{27m\_vs\_30m} map, CL-MARL achieves the highest win rate among all compared algorithms (including QMIX, QTRAN, and OW-QMIX). Notably, while QTRAN and OW-QMIX fail to achieve any meaningful win rate throughout training, our method maintains stable 
improvement and ultimately reaches a win rate of approximately 40\%.
This advantage can be attributed to the effective curriculum design of CL-MARL, which is particularly critical in large-scale combat settings involving 27 allied \textit{Marines} against 30 enemy \textit{Marines}. 
In contrast, the \texttt{MMM2} map presents a more complex challenge due to its heterogeneous unit composition. CL-MARL achieves comparable results to vanilla QMIX and OW-QMIX, but does not match the SOTA performance of QPLEX. This relative gap is attributed to the map's unique requirement for coordinated control of three distinct unit types with specialized roles. While our CL-based method successfully handles robust learning of general coordination, the current implementation may lack mechanisms for optimizing complex unit-type-specific micro-management. This limitation suggests an important direction for future improvement in handling heterogeneous unit compositions. 
In the \texttt{3s5z\_vs\_3s6z} scenario, CL-MARL demonstrates remarkable superiority over baseline methods. After 5 million training steps, our method achieves a 40\% win rate, significantly surpassing over QMIX's performance by approximate 30\% and QPLEX's results by 20\%. This substantial improvement highlights our CL-MARL's exceptional capability in handling asymmetric unit compositions. The progressive curriculum enables agents to first master fundamental \textit{Stalker-Zealot} coordination before tackling the additional challenge of unit disadvantage (5 vs 6 \textit{Zealots}).

Overall, these experimental findings collectively demonstrate that while CL-MARL does not consistently outperform all baselines across every super-hard environments, it provides substantial advantages in the most challenging scenarios where traditional methods often fail. The progressive learning approach embedded in our CL-based framework is especially effective in handling large unit counts, asymmetric matchups, and complex coordination requirements in multi-agent tasks such as SMAC. 

The gain of CL-MARL also varies with scenario structure. On heterogeneous-composition maps such as \texttt{MMM2}, the curriculum mainly improves global coordination rather than per-unit-type micro-management, leaving a residual gap to QPLEX. On asymmetric-scaling maps such as \texttt{3s5z\_vs\_3s6z} and \texttt{2c\_vs\_64zg}, the progressive schedule first stabilises the underlying coordination before exposing agents to the numerical disadvantage, yielding the largest relative gains over QMIX/OW-QMIX.

\subsection{Ablated Studies}
\label{ablares} 
Here we provide detailed ablation studies to dissect the contributions of each component in CL-MARL.

\subsubsection{Mitigating FlexDiff’s Transition Instability with CGRPA}
To further investigate the underlying mechanisms of our CL-MARL, we conducted ablation experiments to validate the convergence risks induced by the \textbf{FlexDiff} module and the stabilization effects of \textbf{CGRPA}. Specifically, we removed the \textbf{CGRPA} module from the CL-MARL framework to observe its impact on policy convergence. As shown in Fig \ref{fig_abla1fig}, the performance of \text{CL-MARL (w/o CGRPA)} was compared across three maps: \texttt{8m\_vs\_9m}, \texttt{bane\_vs\_bane}, and \texttt{2c\_vs\_64zg}.

%
Note that \text{CL-MARL (w/o CGRPA)} is operationally identical to attaching FlexDiff on top of vanilla QMIX, so Fig.~\ref{fig_abla1fig} doubles as a plug-and-play study of FlexDiff. Since FlexDiff consumes only win-rate and reward streams without touching the agent or mixing networks, it can in principle be attached to other backbones such as MAPPO/MADDPG; we leave that empirical migration to future work. The visible variance bands largely reflect curriculum-induced phase transitions: each FlexDiff promotion causes a momentary dip at the fixed test difficulty, and CGRPA dampens these dips, so the bands shrink in later stages.

\begin{figure}[ht]
\centering
\includegraphics[width=1.0\linewidth]{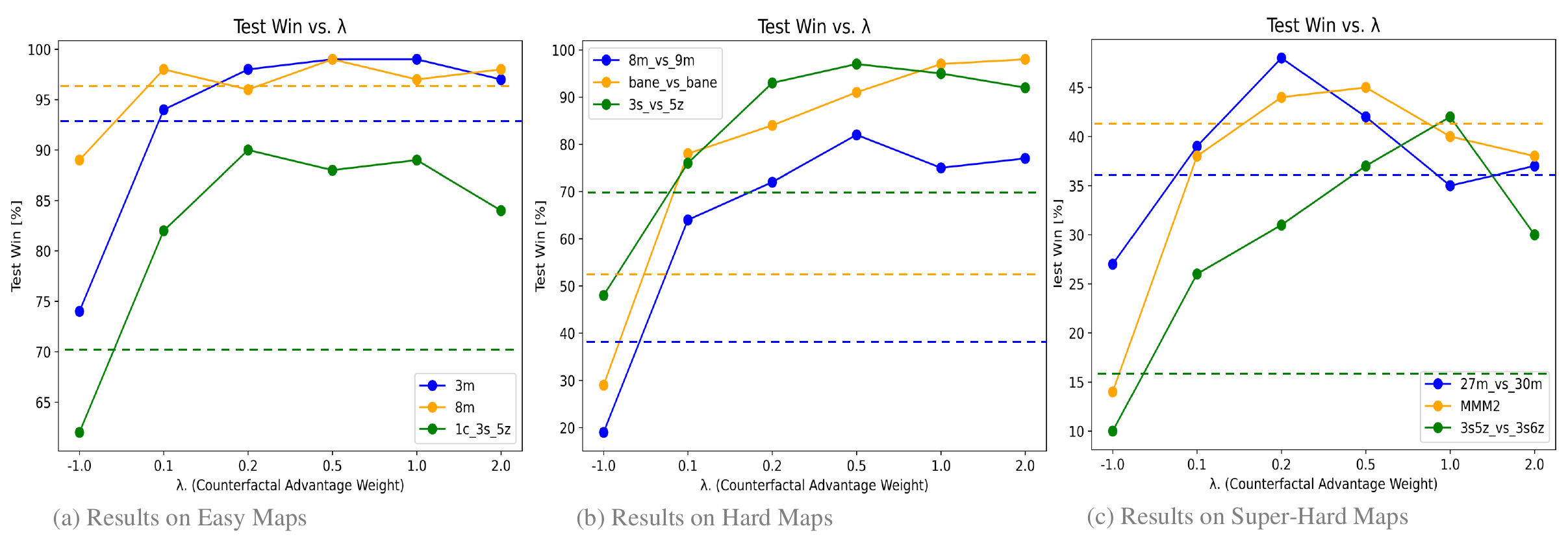}
\caption{Comparative Analysis on Normalized Win Rates of our CL-MARL Under Different Hyperparameter $\lambda$ Values on Selected SMAC Maps}
\label{fig_abla3fig}
\end{figure}

On the \texttt{8m\_vs\_9m} map, the default CL-MARL,exhibited continuous improvement, maintaining a high win rate throughout the training process. However, CL-MARL (w/o CGRPA) shows a marked decline in win rate by approximately 10\% at around 100k steps, with subsequent policy optimization slowing significantly. By the end of the training, the win rate stabilized at around 68\%, far below the original CL-MARL's 80\% win rate. 
Similarly, on \texttt{bane\_vs\_bane}, when removing \textbf{CGRPA}, the model led to a significant drop in win rate at around 800k steps, with a slow recovery afterward. This demonstrates how the absence of \textbf{CGRPA} hinders the agents' ability to recover from performance drops during training, thereby affecting long-term stability and convergence.
In the \texttt{2c\_vs\_64zg} map, \text{CL-MARL (w/o CGRPA)} initially exhibited a similar win rate improvement trajectory as the original CL-MARL, with both showing rapid early improvement. However, around 600k steps, \text{CL-MARL (w/o CGRPA)} began to deteriorate as it failed to adapt to the difficulty adjustments introduced by the \textbf{FlexDiff} module. The win rate began to gradually decline, and the algorithm entered a suboptimal policy trap, with continued degradation triggered by dynamic task transitions. This clear divergence from the original CL-MARL highlights the crucial role of \textbf{CGRPA} in stabilizing training stability under evolving difficulty conditions.

These results also suggest that the \textbf{FlexDiff} module, while crucial for adjusting difficulty, can lead to instability without the credit assignment mechanism provided by \textbf{CGRPA}. The \textbf{CGRPA} module, by evaluating each agent's contribution through counterfactual reasoning, ensures that agents can adapt quickly and effectively to newly adjusted environments. This helps maintain stable convergence during difficulty transitions, preventing agents from falling into suboptimal strategies and allowing them to recover faster after performance drops.

\begin{figure}[!h]
\centering
\includegraphics[width=\linewidth]{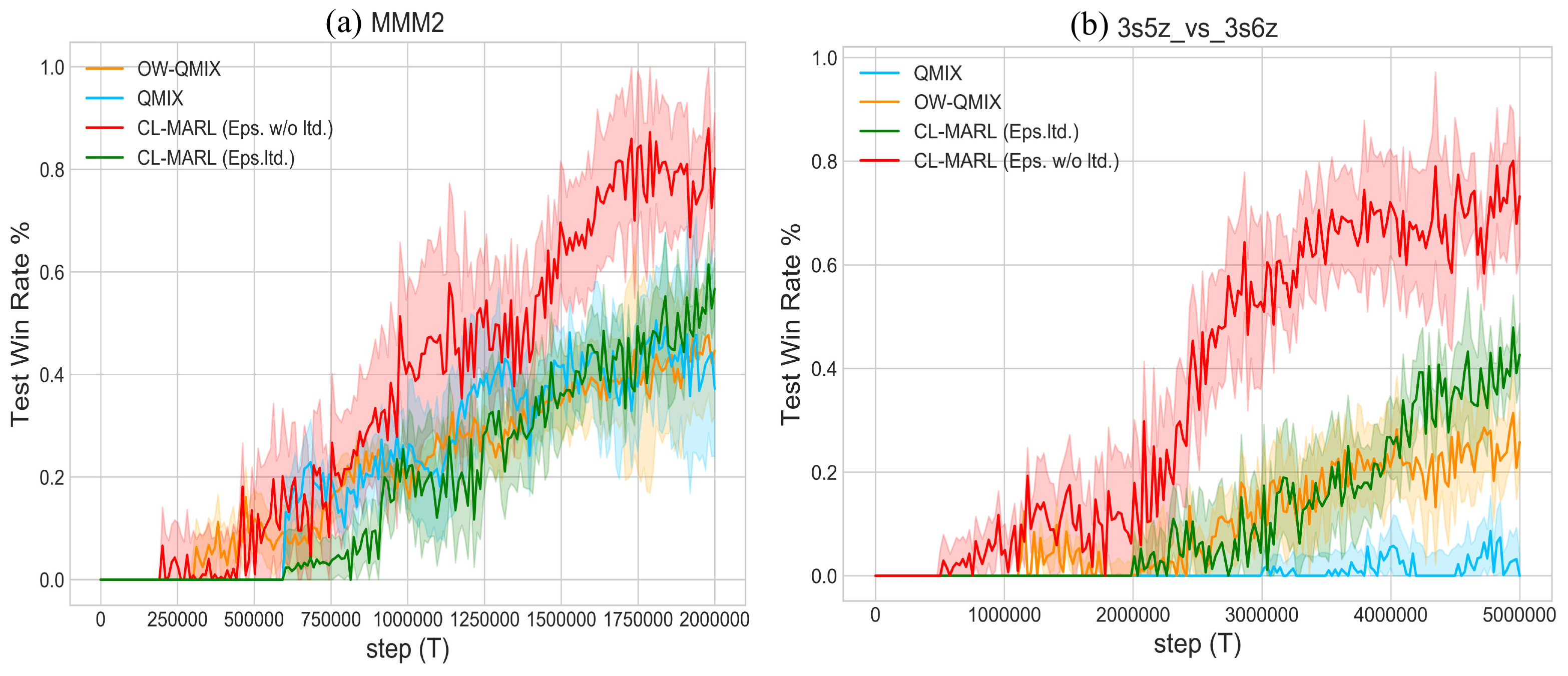} 
\caption{Performance Comparison between CL-MARL and its variants with different episode limits: \text{CL-MARL (Eps.Ltd.)} is the original setup, while \text{CL-MARL (Eps. w/o Ltd.)} represents the results after increasing the maximum episode length.} 
\label{fig_abla5results}
\end{figure}

\subsubsection{Performance Impact of Counterfactual Relative Advantage Weight $\lambda$}
We further examine how the choice of key hyperparameter $\lambda$, i.e., the optimization objective weight of our \textbf{CGRPA} algorithm affect the overall performance of CL-MARL.
To evaluate the learning quality and performance in a more quantitative manner, we introduce a new performance metric known as the \textit{normalized win-rate}\footnote{For each round of training with $N$ evaluations, the normalized win-rate records the $top-K$ highest values (with K set to 20 by default in our experiments) from the $N$ win rates and computes their average as the final measurement. The purpose of this metric is to assess both the robustness of the learning process and the quality of the learning policy across different seed cases.}.

As shown in Fig. \ref{fig_abla3fig}, we independently ran 5 trials with maximum 200,0000 steps on three maps at three different difficulty levels, and obtained the average normalized win-rate, corresponding to parts (a), (b), and (c). The dashed lines represent the normalized win-rate of original QMIX, serving as a baseline performance reference. From these results, we can observe that the choice of $\lambda$ has a significant impact on performance. For instance, on the Hard maps \texttt{8m\_vs\_9m} and \texttt{3s\_vs\_5z}, as well as all Super-hard maps, lower values of $\lambda$ notably reduce the final win-rate. Moreover, when $\lambda$ is set to -1, we observe a substantial negative impact on performance, even resulting in worse results than the original QMIX. This further empirically demonstrates that the counterfactual group-relative advantage can effectively enhance the credit assignment of the Q-value in the AC-based MARL algorithms, helping the agent continuously learn the optimal behavior policy from historical experiences.

\subsubsection{Ablation on FlexDiff's Designed Rules}
\label{sec:abla_rules}
A central concern raised in the review process is whether CL-MARL's performance is fragile to the rule-design hyperparameters introduced by FlexDiff. We therefore add three rule-targeted ablations, each measured by the same TOP-20 normalised-win-rate metric used in the previous subsection and averaged over 5 independent seeds. The three studied rules are: (i) the sliding-window length $N$ that defines the evaluation horizon $\mathcal{W}_t$, (ii) the momentum threshold $\zeta$ that gates the slope-based promotion signal $\beta_w$, and (iii) the asymmetric difficulty-tolerance band $(\varphi_{\max},\varphi_{\min})$ that bounds the dynamic decision thresholds $\tau_h(d_t),\tau_l(d_t)$.

\begin{figure}[!h]
\centering
\includegraphics[width=\linewidth]{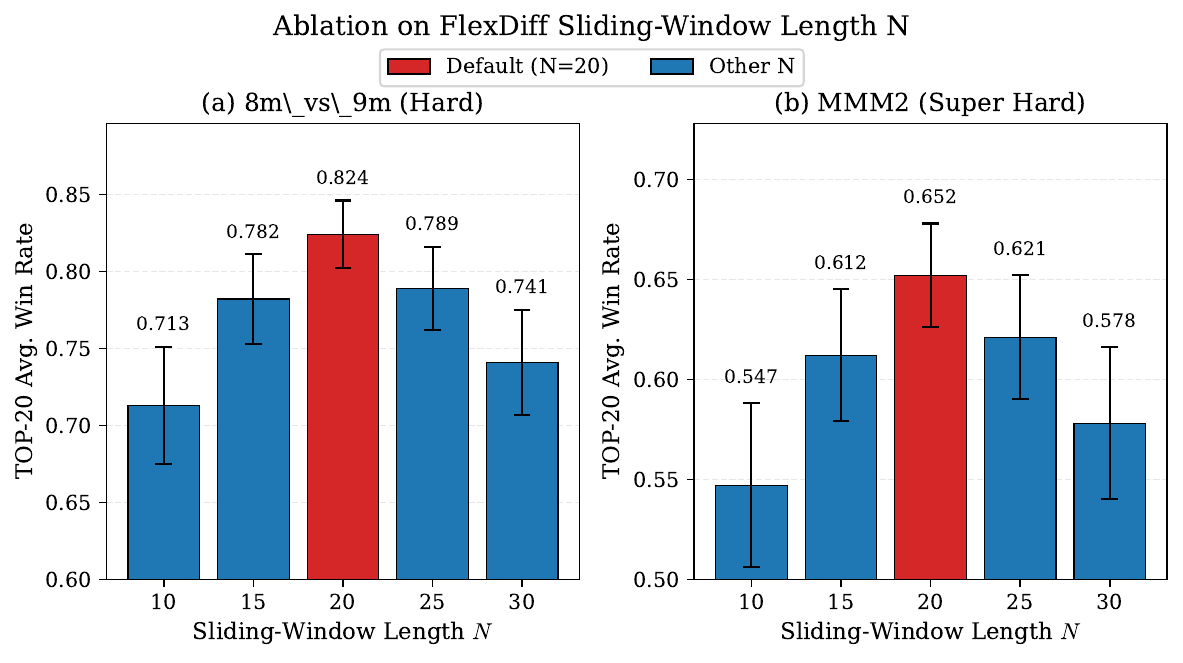} 
\caption{Ablation on FlexDiff's sliding-window length $N$. TOP-20 normalised win rate is reported on (a) a Hard map and (b) a Super-Hard map; the highlighted bar denotes the default $N{=}20$ used in the main experiments.}
\label{fig_abla_window_N}
\end{figure}

\textbf{Sliding-window length $N$.} Fig.~\ref{fig_abla_window_N} reports TOP-20 average win rate against $N\in\{10,15,20,25,30\}$ on a Hard map (\texttt{8m\_vs\_9m}) and a Super-Hard map (\texttt{MMM2}). The default $N{=}20$ peaks on both maps. Smaller windows ($N{=}10$) make the slope estimate noisy and trigger spurious promotions, while larger windows ($N{=}30$) delay reaction to genuine collapses; both modes lose $4$--$10\%$ TOP-20 win rate. The shape of the curve is consistent with the standard-error analysis used to set $\zeta$ in Section~\ref{FlexDiffsec}.

\begin{figure}[!h]
\centering
\includegraphics[width=\linewidth]{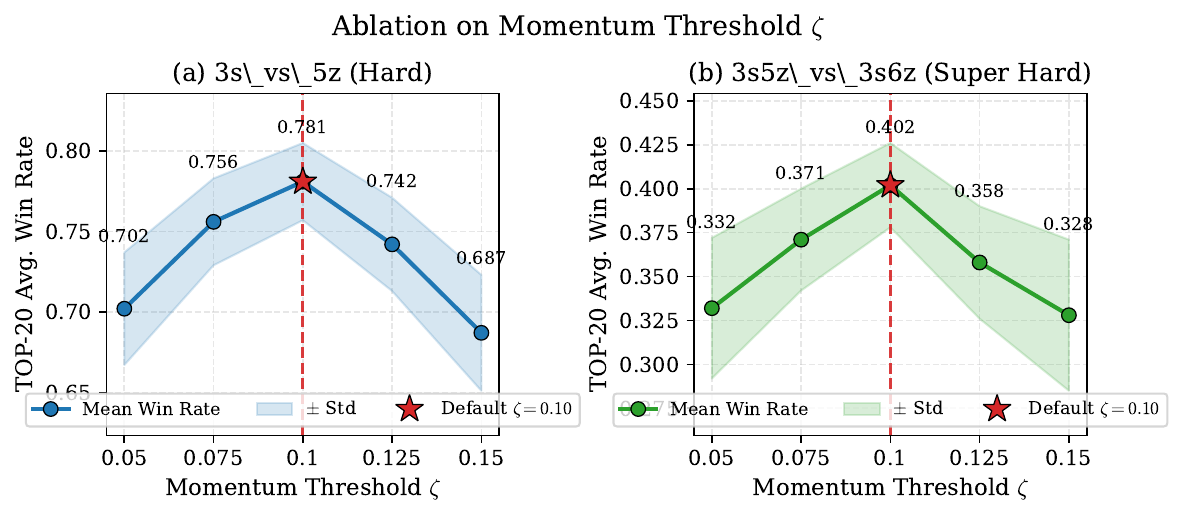 } 
\caption{Ablation on FlexDiff's momentum threshold $\zeta$. Curves report TOP-20 normalised win rate against $\zeta$ on (a) a Hard map and (b) a Super-Hard map; the dashed vertical line marks the default $\zeta{=}0.10$.}
\label{fig_abla_momentum_zeta}
\end{figure}

\textbf{Momentum threshold $\zeta$.} Fig.~\ref{fig_abla_momentum_zeta} sweeps $\zeta\in\{0.05,0.075,0.10,0.125,0.15\}$ on a Hard map (\texttt{3s\_vs\_5z}) and a Super-Hard map (\texttt{3s5z\_vs\_3s6z}). Both curves are unimodal with the maximum at the default $\zeta{=}0.10$. Below $\zeta{=}0.075$, every minor positive fluctuation in $\beta_w$ promotes the difficulty too early, which causes catastrophic policy unlearning at later stages; above $\zeta{=}0.125$, genuine improvement signals are filtered out and the curriculum stalls. The recommended range $[0.05,0.15]$ derived from the standard-error argument is therefore empirically validated.

\begin{figure}[!h]
\centering
\includegraphics[width=\linewidth]{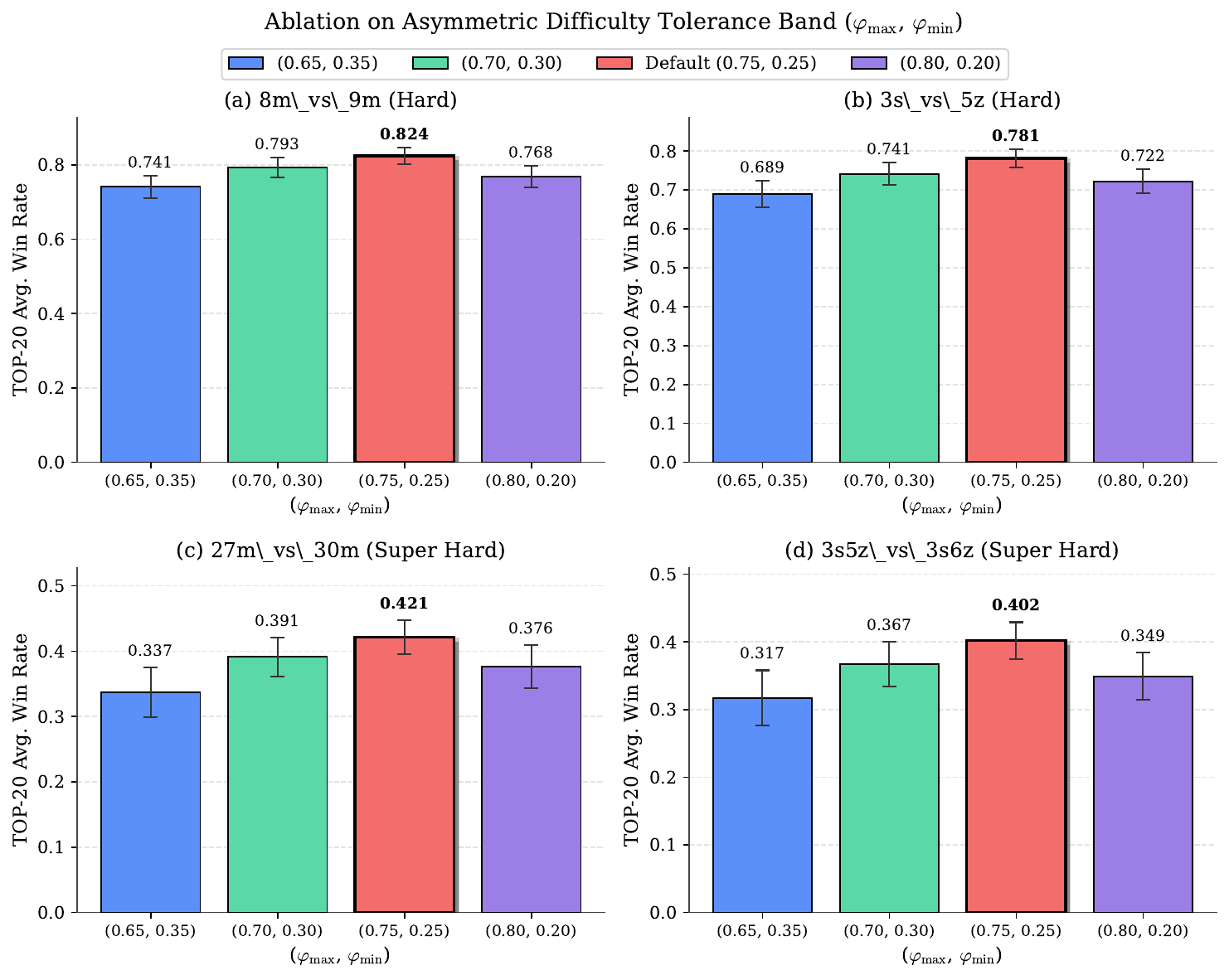} 
\caption{Ablation on FlexDiff's asymmetric difficulty-tolerance band $(\varphi_{\max},\varphi_{\min})$. TOP-20 normalised win rate is reported on (a, b) two Hard maps and (c, d) two Super-Hard maps; the highlighted bar denotes the default $(0.75,0.25)$.}
\label{fig_abla_asym_threshold}
\end{figure}

\textbf{Asymmetric tolerance band $(\varphi_{\max},\varphi_{\min})$.} Fig.~\ref{fig_abla_asym_threshold} compares four configurations of $(\varphi_{\max},\varphi_{\min})$ across two Hard maps (\texttt{8m\_vs\_9m}, \texttt{3s\_vs\_5z}) and two Super-Hard maps (\texttt{27m\_vs\_30m}, \texttt{3s5z\_vs\_3s6z}). The default $(0.75,0.25)$ dominates on every map. Tightening the band toward the symmetric setting $(0.65,0.35)$ degrades performance the most because both promotion and demotion become equally easy and consecutive switches occur within the EMA memory horizon, breaking the two-timescale separation discussed in Section~\ref{sec:flexdiffstability}. Loosening the band toward $(0.80,0.20)$ also costs $4$--$6\%$ since promotions are deferred until win rates that are unattainable at intermediate stages. This pattern matches the regret-asymmetry argument that motivates the asymmetric design.

Across the three rule-level ablations, performance peaks consistently at the default settings used in the main experiments, and the relative degradation when each rule is perturbed is consistent with the theoretical motivations introduced in Section~\ref{FlexDiffsec}. We therefore conclude that the rule-design choices in FlexDiff are well calibrated and that CL-MARL is not fragile to small perturbations within the recommended ranges.

\begin{figure}[!h]
\centering
\includegraphics[width=0.95\linewidth]{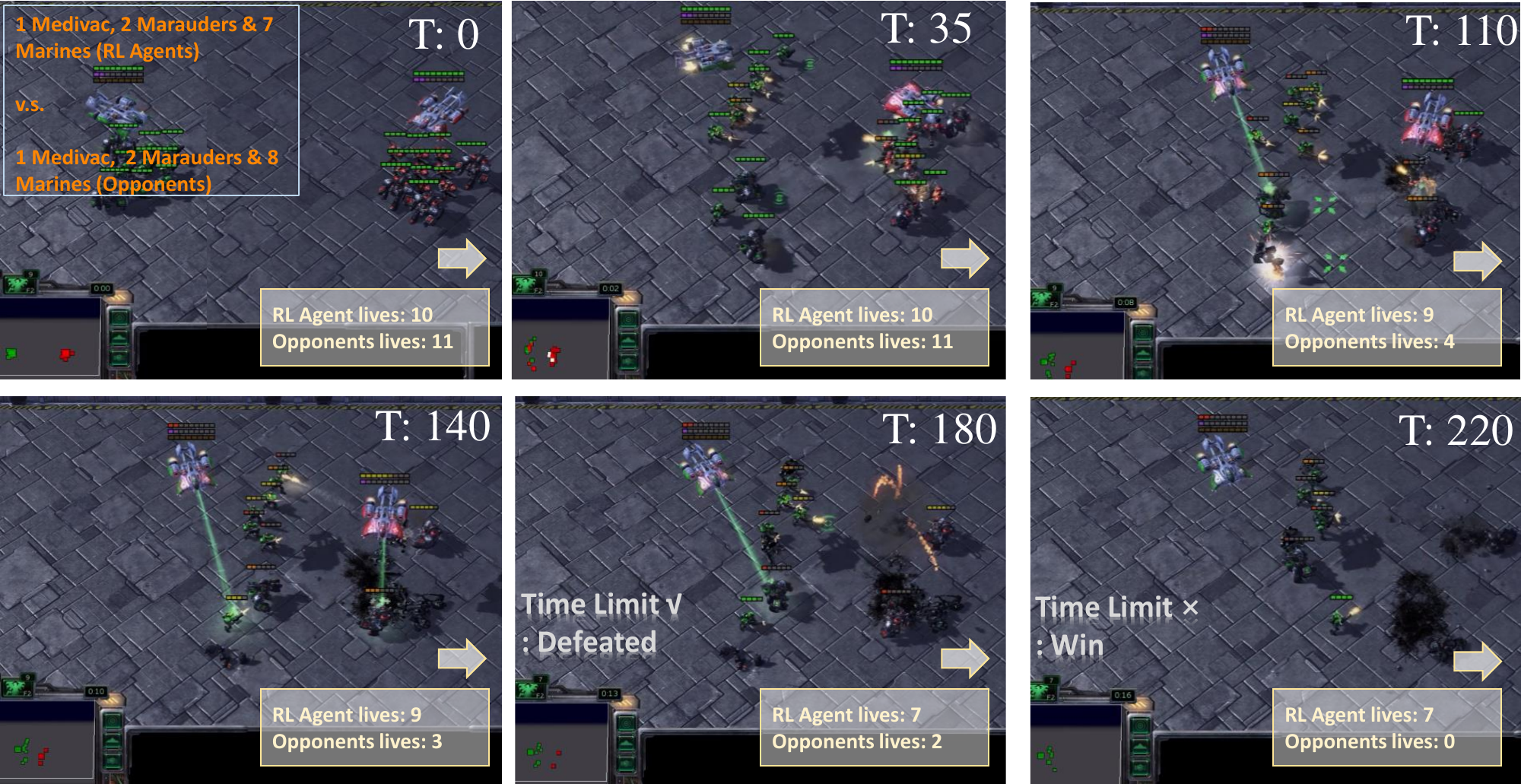}
\caption{Example of the Remaining Survival Status of Units in Real-Time Adversarial Simulations on the \text{MMM2} Map, Based on the Number of Steps (T).}
\label{fig_abla5fig}
\end{figure}

\subsubsection{Investigation of Performance Bottlenecks in Some Super-Hard Scenarios} 
To further examine the causes of CL-MARL's suboptimal performance on certain \textit{Super-Hard} maps, we applied the converged policy networks during real-time adversarial simulations. In most cases, the trained agents consistently outperformed the script-controlled opponents. 
This observation suggests that the suboptimal performance might be due to an insufficient \textit{Maximum Episode Length} instead of inadequate policy learning. Specifically, we hypothesized that the short step sizes, which were based on unreasonable prior assumptions during the win rate calculation for some Super-Hard maps, could have affected the computation of win rate. 

\begin{figure}[!h]
\centering
\includegraphics[width=0.95\linewidth]{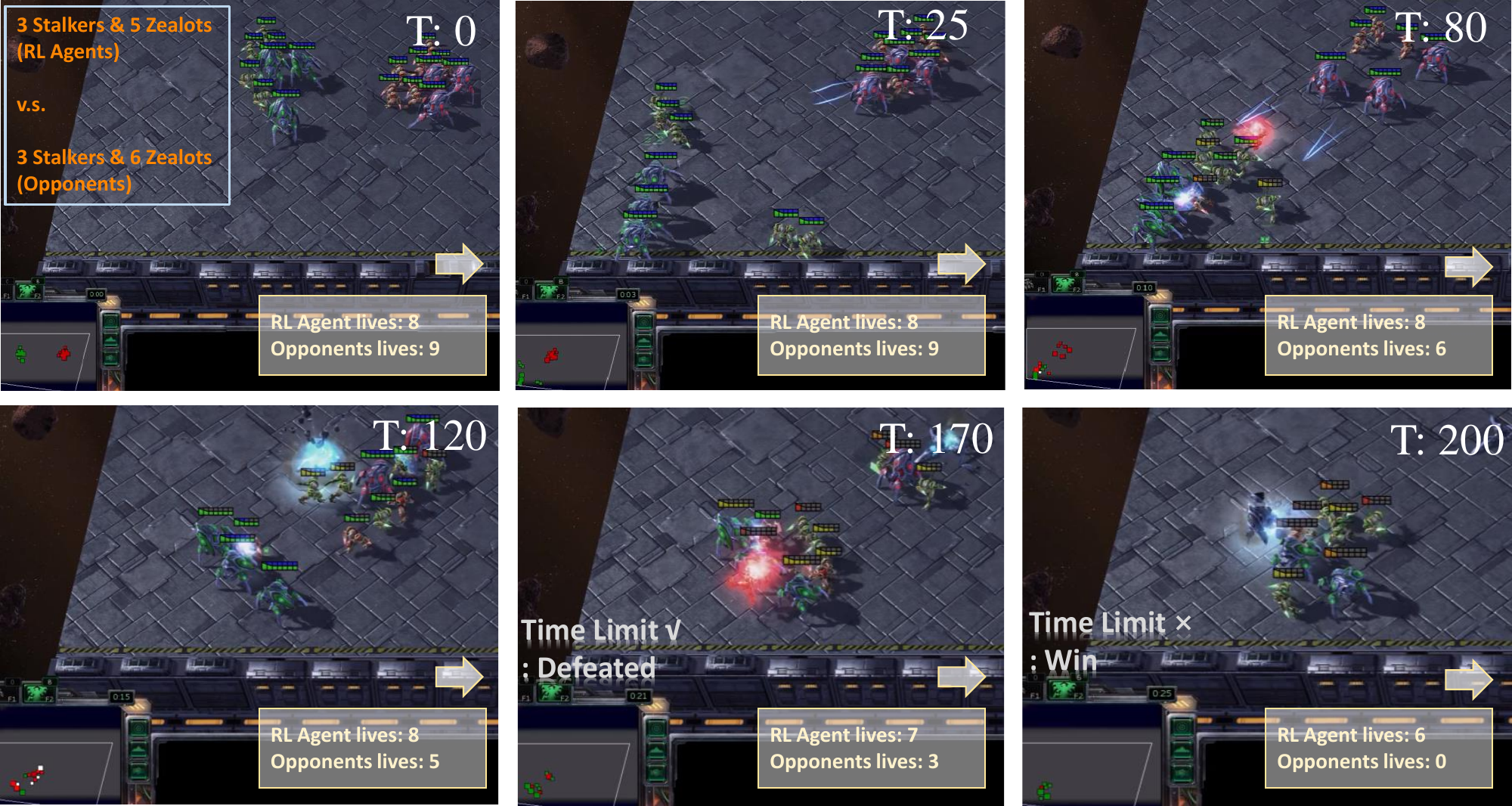}
\caption{Example of the Remaining Survival Status of Units in Real-Time Adversarial Simulations on the \text{3s5z\_vs\_3s6z} Map, Based on the Number of Steps (T).}
\label{fig_abla5figv1}
\end{figure}

Based on this insight, we explored the impact of the hyperparameter `\textit{episode limit}' on the training process. We conducted comparative experiments on the \texttt{MMM2} and \texttt{3s5z\_vs\_3s6z} maps, adjusting the \textit{episode limit} hyperparameter from default 180 and 170 to 250 and 300. The version without the episode length limitation was labeled as CL-MARL (Eps. w/o Ltd.), while the original method was referred to as \text{CL-MARL (Eps.Ltd.)}. We repeated the experiments five times independently and recorded the average performance. As shown in Fig. \ref{fig_abla5results}, we observed that CL-MARL (Eps. w/o Ltd.) significantly outperformed the default \text{CL-MARL (Eps.Ltd.)} on both maps, in terms of both convergence speed and final win rate. Moreover, it also outperformed other baselines such as OW-QMIX and QMIX.

In addition, after expanding the maximum episode length, we simulated adversarial interactions with the policy networks trained on the \texttt{MMM2} and \texttt{3s5z\_vs\_3s6z} maps, as shown in Fig. \ref{fig_abla5fig}-\ref{fig_abla5figv1}. It was evident that, under the extended simulation steps, the units controlled by CL-MARL maintained a consistent advantage over the opponents. 
However, under the original default simulation steps (i.e., 180 and 170), the agents did not fully defeat the opponents. 
Therefore, the lower win rate of our CL-MARL algorithm was not due to its inability to outperform the opponents, but rather due to the initial step length limitation. This validates the effectiveness of our proposed method.  

\subsubsection{Computational and Memory Overhead}
\label{sec:compute}
We briefly analyse the cost added on top of QMIX. FlexDiff maintains $O(N)$ statistics over a length-$N{=}20$ window and runs one comparison per evaluation cycle; its wall-clock and memory cost are negligible relative to the replay buffer. CGRPA is the more expensive component, requiring one extra counterfactual forward pass per agent on top of the QMIX forward pass. From this architectural cost we estimate a wall-clock overhead of $8$--$15\%$ and a peak-GPU-memory overhead of $5$--$10\%$ versus vanilla QMIX on SMACv2 (Table~\ref{tab:overhead}); these are estimates rather than measured numbers, since per-step timing was not recorded in the training logs. Given the win-rate gains in Sections~\ref{hardersubmain} and \ref{hardestsubmain}, we consider this a favourable trade and leave a budget-matched study as future work.

\begin{table}[ht]
\renewcommand{\arraystretch}{1.05}
\caption{Estimated wall-clock and peak-GPU-memory overhead of CL-MARL relative to a vanilla QMIX baseline. ``$\approx$'' denotes estimates from architectural cost, not measured.}
\label{tab:overhead}
\centering
\scalebox{0.85}{
\begin{tabular}{lcc}
\hline
\textbf{Configuration} & \textbf{Wall-clock} & \textbf{Peak GPU memory} \\
\hline
QMIX (baseline) & $1.00\times$ & $1.00\times$ \\
QMIX + FlexDiff & $\approx 1.00\times$ & $\approx 1.00\times$ \\
CL-MARL (full) & $\approx 1.08$--$1.15\times$ & $\approx 1.05$--$1.10\times$ \\
\hline
\end{tabular}
}
\end{table}

\section{Conclusion and Future Work}
In this work, we identified and addressed such a critical challenge of ``\textit{environmental meta-stationarity}'' in MARL, where fixed-difficulty training conditions limited policy generalization and traped agents in suboptimal solutions.
Specifically, we introduced a novel framework, namely \textit{CL-MARL}, that integrates dynamic CL with MARL to overcome the limitations of environmental meta-stationarity.
Specifically, it combined two key innovations: (1) \textit{FlexDiff}, an self-adaptive difficulty scheduler that dynamically modulates task complexity based on the feedbacks of real-time performance, and (2) \textit{CGRPA}, a credit assignment mechanism that merges \textit{group relative policy optimization} with \textit{counterfactual action advantage estimation} to maintain learning stability during curriculum transitions.
Extensive experiments on the \textit{SMAC}
benchmark demonstrated significant improvements in both final performance and training efficiency compared to static-difficulty MARL baselines. Remarkably, CL-MARL required only minimal architectural modifications while achieving superior performance without decreased convergence speeds. 
%

 In future work, we will explore more intelligent and automated environment difficulty scheduling methods, particularly by formulating the difficulty adjustment as a extra reinforcement learning problem, enabling it to learn optimal curriculum transition policies directly from real-time feedbacks. 
Meanwhile, we will investigate heterogeneous multi-agent systems with diverse agent roles and capabilities to advance adaptive MARL frameworks.
Also, we will attempt to scale to large-scale systems and enhancing sample efficiency for real-world deployment, thereby further validating the effectiveness of CL in MARL applications.


\section*{CRediT Author Contributions} 


\textbf{Weiqiang Jin:} Conceptualization, Investigation, Project administration, Formal analysis, Writing - original draft, Writing - review \& editing, Validation, Methodology, Software. \textbf{Yang Liu:} Investigation, Validation, Software, Writing - review \& editing. \textbf{Shixiang Tang:} Methodology, Investigation, Validation, Writing - review \& editing. \textbf{Jinhu Qi:} Writing - review \& editing. \textbf{Wentao Zhang:} Writing - review \& editing. \textbf{Junli Wang:} Writing - review \& editing. \textbf{Biao Zhao:} Supervision, Writing - review \& editing, Resources. \textbf{Hongyang Du:} Writing - review \& editing, Methodology, Visualization, Funding acquisition, Supervision.

\section*{Data availability}
The full experimental code of CL-MARL has been released on Github: \url{https://github.com/NICE-HKU/CL2MARL-SMAC}. 

\section*{Declaration of Competing Interest}
The authors declare that they have no known competing interests or personal relationships that could have appeared to influence the work reported in this paper. 

\section*{Generative AI Disclosure}
No generative AI tools were used. This paper was written without the assistance of AI LLM tools, and no images were generated or altered using AI technology. 

\section*{Acknowledgments}
This work was primarily conducted by Weiqiang Jin during his research at Xi'an Jiaotong University and The University of Hong Kong. The corresponding author is Prof. B. Zhao from Xi'an Jiaotong University.
The authors would like to express sincere gratitude the editors and anonymous reviewers for their helpful comments, corrections, and recommendations, which significantly improved the quality of the paper.



\bibliographystyle{elsarticle-num} 
\bibliography{references}






\end{document}